\documentclass[10pt,twocolumn,letterpaper]{article}

\usepackage[pagenumbers]{templates/cvpr2024/cvpr} 

\usepackage{graphicx}
\usepackage{amsmath}
\usepackage{amssymb}
\usepackage{booktabs}
\usepackage[accsupp]{axessibility}

\usepackage{lipsum} 
\usepackage{enumitem} 
\usepackage{tikz}
\usepackage{bm} 
\usepackage{array} 
\usepackage{multirow}
\usepackage{dirtree}
\usepackage{tabularx}

\usepackage{xfrac}

\usepackage{svg}
\usepackage{arydshln} 

\usepackage{fontawesome}
\usepackage{pifont} 

\usepackage{tikzit}

\definecolor{redpipeline}{HTML}{820263} 
\definecolor{orangepipeline}{HTML}{d90368} 
\definecolor{pinkpipeline}{HTML}{fb8b24} 

\tikzstyle{bigblock}=[fill=white, draw=redpipeline, shape=rectangle, minimum width=6.375cm, minimum height=2.625cm]
\tikzstyle{tbigblock}=[fill=white, draw=orangepipeline, shape=rectangle, minimum width=6.375cm, minimum height=2.625cm] 
\tikzstyle{ttbigblock}=[fill=white, draw=pinkpipeline, shape=rectangle, minimum width=6.375cm, minimum height=2.625cm] 
\tikzstyle{square}=[fill=white, draw=black, shape=rectangle, minimum width=1.5cm, minimum height=1.5cm, opacity=.125, rounded corners]

\newlength{\pipelinelw}
\tikzstyle{arrow}=[draw=black, fill=none, ->,line width=\pipelinelw]
\tikzstyle{tarrow}=[draw=black, ->, opacity=.5,line width=\pipelinelw]
\tikzstyle{tline}=[-, draw=orangepipeline,line width=\pipelinelw] 
\tikzstyle{line}=[-,line width=\pipelinelw]
\tikzstyle{ttarrow}=[->, draw=black, opacity=0.25,line width=\pipelinelw]
\tikzstyle{ttline}=[-, draw=pinkpipeline,line width=\pipelinelw] 
\tikzstyle{darrow}=[<->, draw=black,line width=\pipelinelw]
\usetikzlibrary{positioning,calc,arrows.meta,intersections,through}
\usetikzlibrary{calc,intersections,through,backgrounds}

\usepackage{booktabs}
\renewcommand{\arraystretch}{1.1}

\usepackage{pgfplots}
\pgfplotsset{compat=1.18}

\usepackage{caption} 
\usepackage{subcaption} 
\usepackage{style}

\definecolor{cvprblue}{rgb}{0.21,0.49,0.74}
\usepackage[pagebackref,breaklinks,colorlinks,citecolor=cvprblue]{hyperref}

\usepackage[capitalize]{cleveref}
\crefname{section}{Sec.}{Secs.}
\Crefname{section}{Section}{Sections}
\Crefname{table}{Table}{Tables}
\crefname{table}{Tab.}{Tabs.}

\def\paperID{291} 
\def\confName{CVPR}
\def\confYear{2024}

\title{Learnable Earth Parser: Discovering 3D Prototypes in Aerial Scans}

\author{Romain Loiseau\textsuperscript{1, 2}\\{{\tt\footnotesize romain.loiseau@enpc.fr}}
\and
Elliot Vincent\textsuperscript{1,3}\\{{\tt\footnotesize elliot.vincent@enpc.fr}}
\and
Mathieu Aubry\textsuperscript{1}\\{{\tt\footnotesize mathieu.aubry@enpc.fr}}
\and
Loic Landrieu\textsuperscript{1, 2}\\{{\tt\footnotesize loic.landrieu@enpc.fr}}
\and
{{\textsuperscript{1} LIGM, Ecole des Ponts, Univ Gustave Eiffel, CNRS, France}}\\
{{\textsuperscript{2} Univ Gustave Eiffel, IGN, ENSG, LASTIG, France}}\\
{{\textsuperscript{3} INRIA Paris, France}}
}

\begin{document}
\maketitle

\begin{abstract}
We propose an unsupervised method for parsing large 3D scans of real-world scenes with easily-interpretable shapes. This work aims to provide a practical tool for analyzing 3D scenes in the context of aerial surveying and mapping, without the need for user annotations. Our approach is based on a probabilistic reconstruction model that decomposes an input 3D point cloud into a small set of learned prototypical 3D shapes. The resulting reconstruction is visually interpretable and can be used to perform unsupervised instance and low-shot semantic segmentation of complex scenes. We demonstrate the usefulness of our model on a novel dataset of seven large aerial LiDAR scans from diverse real-world scenarios. Our approach outperforms state-of-the-art unsupervised methods in terms of decomposition accuracy while remaining visually interpretable. Our code and dataset are available at \url{https://romainloiseau.fr/learnable-earth-parser/}.
\end{abstract}
\section{Introduction}\label{sec:intro}

\begin{figure}[t]
    \centering
    \input{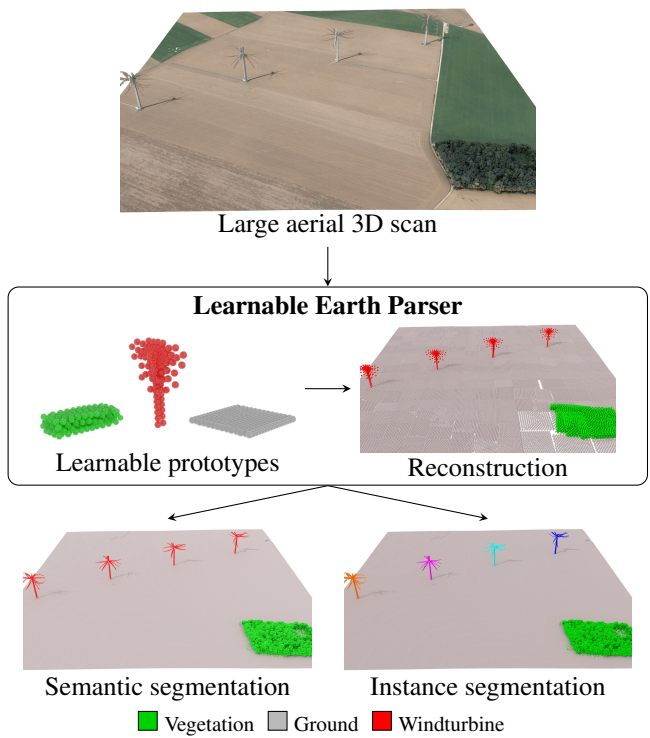}
    \caption{\textbf{Learnable Earth Parser.}~
    Our unsupervised method takes large aerial 3D scans as input and model them with a small set of learned 3D prototypes. Our approach is trained without annotation and produce legible decompositions of complex scenes, which can be used for semantic and instance segmentation.}
    \label{fig:teaser}
    \vspace{\vspacevalue em}
\end{figure}

\begin{figure*}[t]
    \centering
    \ctikzfig{figures/pipeline}{.975}
    \vspace{-1.5em}
    \caption{\textbf{Method Overview.}~
    Our model approximates an input point cloud $\bX$ with $S$ slot models. Each slot maps $\bX$ to an affine 3D deformation $\cT_s(\bX)$, a slot activation probability $\alpha_s$, and the joint probabilities $\beta_s^1, \cdots, \beta_s^K$ of the slot being activated and choosing one of the $K$ learnable prototype point clouds $\proto^1, \cdots, \proto^K$. The output $\cM_s(\bX)$ of an activated slot $s$ is obtained by applying the transformation $\cT_s(\bX)$ to its most likely prototype. Non-activated slots do not contribute to the output.}
    \label{fig:pipeline}
    \vspace{\vspacevalue em}
\end{figure*}

Modern aerial 3D scanning technologies open up unprecedented opportunities for environmental monitoring and economic intelligence. However, their practical use remains challenging due to the complexity of {real-}world scenes, the diversity of usage scenarios, and the difficulty of annotation. Therefore, our aim is to develop an approach that could help perform diverse tasks---from counting trees in a forest or identifying the various components of a factory to measuring the surface of greenhouses or monitor urban growth---all without human supervision.

To do so, we address two important limitations of existing 3D deep learning methods. First, they are often primarily designed, trained, and tested on synthetic~\cite{modelnet,shapenet,shapenetsem,McCormac2017Scene} or highly curated data~\cite{armeni20163d,kitti360,tankandtemples}, which fail to capture the endless variability of the real world. Moreover, they often assume that annotations are available for tasks of interest. 
Second, even unsupervised approaches~\cite{alliegro2021joint,Zhang_2021_ICCV} often rely on learning abstract feature representations, making them difficult to interpret~\cite{zhang2021survey}. Although some work has attempted to decompose 3D shapes into meaningful components without supervision~\cite{tulsiani2017learning,deprelle2019learning,paschalidou2019superquadrics,loiseau2021representing}, they were all designed on simple synthetic shapes and none generalizes to real data.

To overcome these limitations, we present the \METHODNAME, an unsupervised deep learning method designed to decompose large-scale 3D point clouds into interpretable parts. Our model learns a small set of 3D prototypical shapes that are selected, positioned, rotated, and resized to reconstruct an input point cloud. We introduce a novel probabilistic formulation that enables the design of a reconstruction loss for  learning jointly the 3D prototypes, but also to select and position them.

To evaluate the effectiveness of our approach, we created a new open-access dataset consisting of 7 aerial LiDAR scans, covering $7.7$km$^2$ and containing $98$ million 3D points with annotations in diverse urban and natural environments. Our results demonstrate that the \METHODNAME \ learns decompositions superior to traditional and deep learning baselines, leading to convincing performance for semantic and instance segmentation, as shown in \figref{fig:teaser}. We believe that our contributions provide researchers and practitioners with new tools and resources to tackle the challenges of real-world 3D data.
\section{Related work}\label{sec:related_work}

\paragraph{}Our proposed unsupervised method uses point cloud reconstruction as a proxy to learn to decompose large aerial point clouds and is evaluated on a novel and diverse dataset of 3D scans. In the following, we briefly present related works for primitive-based point cloud decomposition, automatic decomposition of LiDAR data, and an overview of existing aerial LiDAR datasets.

\paragraph{Primitive-based point cloud decomposition.}Modeling shapes as a set of primitives such as generalized cylinders~\cite{binford1975visual} or superquadrics~\cite{barr1981superquadrics} has a rich history in vision and graphics~\cite{kaiser2019survey}. Classical applications include reverse engineering~\cite{benkHo2001algorithms}, shape completion~\cite{schnabel2009completion,sung2015data}, and shape editing~\cite{gal2009iwires}. A variety of methods have been developed to find primitives in unstructured 3D scenes, including seed growing techniques~\cite{leonardis1995segmentation,leonardis1997superquadrics}, genetic algorithms~\cite{chen1999quadric}, approaches~\cite{gotardo2003range,schnabel2007efficient,li2011globfit,ramamonjisoa2022monteboxfinder} based on RANSAC~\cite{fischler1981random}, and probabilistic methods~\cite{liu2022robust,wu2022primitive}. 

For this problem, like for many others in computer vision, deep learning has become the dominant paradigm. However,  supervised methods~\cite{zou20173d,li2019supervised} are limited by the availability of annotated datasets. Following the seminal work of Tulsiani et al.~\cite{tulsiani2017learning}, unsupervised approaches that simply rely on a reconstruction loss to learn primitive decomposition are the most common and the most closely related to our work~\cite{paschalidou2019superquadrics,paschalidou2021neural}. An important challenge for these approaches is to model a variable number of primitives. This has been addressed using recurrent networks~\cite{zou20173d,li2017grass,sharma2018csgnet}, capsule networks~\cite{zhao20193d}, reinforcement learning strategies~\cite{tulsiani2017learning} or, most similar to our approach, computing the Chamfer distance based on a probabilistic model~\cite{paschalidou2019superquadrics}. Our work is also related to approaches that learn prototypical shapes instead of being restricted to a predefined family of parametric primitives~\cite{deprelle2019learning,loiseau2021representing}.

However, most of these methods are designed, trained and evaluated on well-curated synthetic object datasets, such as ModelNet~\cite{modelnet}, ShapeNet~\cite{shapenet}, or D-FAUST~\cite{bogo2017dynamic}, and are typically designed to handle single objects from known categories. In contrast, our approach can handle complex scenes composed of many objects with significant variety.

\paragraph{Decomposition of LiDAR scans.}Automatically decomposing large LiDAR scans poses unique challenges due to their size and diversity \cite{xia2020geometric}. Some previous approaches use simple shape primitives, such as lines~\cite{guo2016improved,cui2019automatic}, planes~\cite{nan2017polyfit,ge2019multi,guinard2019piecewise}, or volumes~\cite{li2016manhattan}, but these may not be flexible enough to capture the complexity of real-world scans. Other approaches are designed for specific object classes, like trees~\cite{mei20173d} or buildings~\cite{lafarge2006automatic,rs14092254}, but they are limited in their ability to represent a wide range of shapes. In contrast, our \METHODNAME~overcomes these limitations by learning ad-hoc prototypes for each new scene, ensuring both expressivity and adaptability. LiDAR data are often treated as digital elevation models, \ie images with pixel elevations ~\cite{haala1999extraction,lodha2006aerial, guiotte2020semantic}. Thus, our work is related to image-based primitive prediction~\cite{roberts1963machine,gupta2010blocks,kluger2021cuboids,paschalidou2020learning} and unsupervised multi-object image segmentation~\cite{locatello2020object,smirnov_marionette_2021,monnier2021unsupervised,yuUnsupervisedDiscoveryObject2021}. However, 3D point clouds have higher precision and can better represent multi-layered structures such as forest areas.

\paragraph{Aerial LiDAR datasets.}The increased availability of aerial LiDAR technology has led to the multiplication of open datasets~\cite{niemeyer2014contextual,ye2020lasdu,gao2021sum,zolanvari2019dublincity} of varying sizes from $1$ to $10$ km$^2$~\cite{varney2020dales,singer2021dales}. However, these scans are limited to dense urban environments and do not capture the challenge of modeling diverse terrains. Some specialized datasets focus on forested areas~\cite{Kalinicheva_2022_CVPR,weinstein2019individual}. Our proposed dataset is of similar scale, spanning $7.7$km$^2$, but covers a variety of urban, natural, and rural scenes, making it more representative of the diversity of possible usage scenarios.
\section{Method}\label{sec:method}

Our goal is to learn to break down a point cloud into simpler and more easily understandable components. To achieve this, we propose an \emph{analysis-by-synthesis} approach where we train a highly-constrained model $\cM$ to approximate a point cloud as a combination of learned 3D shapes.

\subsection{Probabilistic Scene Reconstruction Model}

As illustrated in \figref{fig:pipeline}, our model first selects up to $S$ shapes from $K$ learnable 3D shapes, then positions and deforms them to best approximate an input point cloud $\bX$. We propose a probabilistic formulation of the selection process, which can be seen as an extension of the model of Paschalidou~\etal~\cite{paschalidou2019superquadrics} to multiple free-form shapes instead of a single parametric family.

\paragraph{Learnable shape prototypes.} Following Loiseau~\etal~\cite{loiseau2021representing}, we define $K$ point clouds $\proto^1,\cdots,\proto^K$ that we refer to as \emph{prototypes}. Each prototype is meant to represent a single instance of a recurring 3D structure in the considered scene. The points' coordinates are free parameters of the model and learned directly.

\paragraph{Scene reconstruction model.} 
Our full model $\cM$ is the combination of $S$ reconstruction models $\cM_s$, which we refer to as \emph{slots} in analogy to the Slot Attention approach \cite{locatello2020object}. 
Each slot contributes to the final reconstruction only if it is activated. Slot activation is determined by a binary variable $a_s$: $\cM_s$ is activated if and only if $a_s=1$. The output of $\cM(\bX)$ is the combination of the reconstructions from all activated slots:
\begin{align}
    \cM(\bX)=\bigcup_{
    \substack{s=1\cdots S\\a_s=1}
    } \cM_s(\bX)~.
\end{align}
Each slot model outputs a point cloud which is the deformation of one learnable prototype chosen from $\proto^1,\cdots,\proto^K$. We associate to each slot $s$ a network $\translater_s$ which maps $\bX$ to an affine transformation in 3D space $\translater_s(\bX)$.  The output $\cM_s(\bX)$ of slot $s$ is determined by a variable $b_s \in \{1,\cdots,K\}$. If $b_s=k$, then the output of $\cM_s(\bX)$ is $\bY_s^k$, the result of applying the transformation $\translater_s(\bX)$ to the prototype $\proto^k$:
\begin{align}
   \bY_s^k=  \translater_s(\bX)[\proto^k]~.
\end{align}
Please note that $\bY_s^k$ is a function of $\bX$. However, to keep our notations simple, we omit this dependence.

\paragraph{Probabilistic modeling.} 
We make our reconstruction model probabilistic by modeling $a$ and $b$ as random variables following (multi-)Bernoulli distributions. We call $\alpha_s$ the probability that slot $s$ is activated and $\beta_s^k$ the probability that it is activated and selects the prototype $k$:
\begin{align}
   p(a_s=1) = \alpha_s~~~,~~~
   p(a_s=1, b_s=k) = \beta_s^k~.
\end{align}
For each slot $s$, we predict the vector $(1-\alpha_s,\beta^1_s,...\beta^K_s)$ with a neural network taking the point cloud $\bX$ as input and finishing with a softmax layer. Again, we don't write the dependency of the $\alpha_s$ and $\beta_s$ on $\bX$ explicitly to simplify the notations. The complete model $\cM(\bX)$ and the slots models $\cM_s(\bX)$ can now be seen as random variables, producing different potential reconstructions with probabilities given by $\alpha$ and $\beta$. During inference, we consider only slots with $\alpha_s>0.5$  and select the prototype with highest $\beta_s^k$. However, during training, we compute all reconstructions $\bY_s^k$.

\begin{figure}
    \centering
    \input{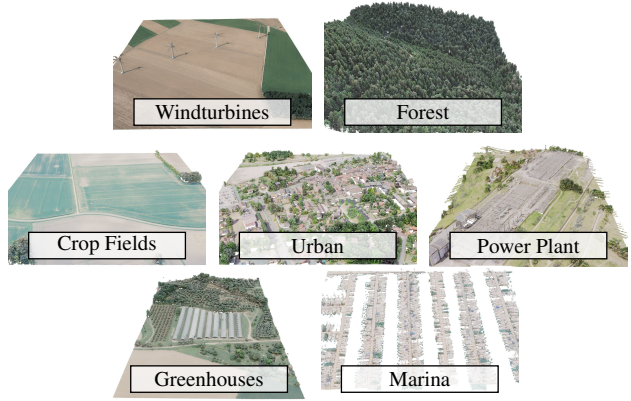}
    \caption{\textbf{\DATASETNAME.}~Our dataset contains 7 scenes representing various urban and natural environments acquired by aerial LiDAR. The illustration of the power plant and the greenhouses display the complete scenes, while other ones display a subset of each scene (between $25$ and $50$\% of the total area).}
    \label{fig:lidarhd:scenes}
    \vspace{\vspacevalue em}
\end{figure}

\subsection{Training Losses}
\label{sec:losses}

Given a large 3D scene, we train our model by sampling square patches $\bX$ from the scene. For each batch of patches, we minimize a loss composed of a reconstruction loss $\cL_\text{rec}$ and several regularization terms $\cL_\text{reg}$ implementing different priors:
\begin{align}
    \cL(\cM) = 
    {\mathbb{E}_\bX}
    \left[ 
        \cL_\text{rec}\left(\cM, \bX\right)
    \right] 
    +
    \cL_\text{reg}\left(\cM\right)~.
\end{align}

\paragraph{Reconstruction loss.} We define the reconstruction loss $\cL_\text{rec}$ as the sum of two losses:
\begin{align}
    \cL_\text{rec}(\cM, \bX) = \cL_{\YfitsX}(\cM, \bX) + \cL_{\XfitsY}(\cM, \bX)~.
\end{align}
$\cL_{\YfitsX}$ encourages likely reconstructions of $\cM(\bX)$ to accurately approximate $\bX$, and $\cL_{\XfitsY}$ ensures coverage, i.e., that each point of $\bX$ is well-reconstructed by at least one activated model. We define each term using the asymmetric Chamfer distance $d$ between two point clouds $\bX$ and $\bY$:
\begin{align} \label{eq:chamfer}
    d(\bX, \bY) = &\frac{1}{|\bX|}\sum_{x\in\bX}\min_{y\in\bY}||x-y||_2^2~,
\end{align}
with $|.|$ the number of points in a point cloud. We write $d(x,\bY)$ the distance between the point $x$ and its closest point in $\bY$.
 
We define $\cL_{\YfitsX}$ as the average over all slots $s$ of the expected distance between $\cM_s(\bX)$ and $\bX$:
\begin{align}
    \cL_{\YfitsX}(\cM, \bX)
    &= \frac{1}{S} 
    \sum_{s=1}^S 
    \mathbb{E}_{a_s,b_s} \left[
        d\left(\cM_s(\bX),\bX\right)
    \right]
    \\
    &=\frac{1}{S}\sum_{s=1}^S\sum_{k=1}^K{\beta_{s}^{k}}{ d\left(\bY_s^k,\bX\right)}~.
    \label{eq:YfitsX}
\end{align}
Conversely, we define $\cL_{\XfitsY}$ as the average over all points $x$ of $\bX$ of the 
expected distance between $x$ and its closest point in the reconstruction:
\begin{align}
    \!\cL_{\XfitsY}(\cM, \bX) = \frac1{|\bX|}\sum_{x \in \bX}
     \mathbb{E}_{a,b}\left[ \min_{s \mid a_s=1} 
        d\left(x,\cM_s(\bX)\right)
     \right]~.
\end{align}
Following the ideas of  Paschalidou~\etal~\cite{paschalidou2019superquadrics}, we first define $\Delta(x,s)$ as the expected distance between $x$ and $\cM_s(\bX)$ conditionally to the slot $s$ being activated:
\begin{align}
    \Delta(x,s)
    &= \mathbb{E}_{b_s \mid a_s=1} \left[d\left(x,\cM_s(\bX)\right)\right]\\
    &= \frac1{\alpha_s}
    \sum_{k=1}^K \beta_s^k d\left(x,\bY_s^k\right)~.
\end{align}
Next, we compute for each point $x$ a permutation $\sigma_x$ of $[1,S]$ such that $\Delta(x,\sigma_x(s))$ is non-decreasing, i.e.:
\begin{align}
    \Delta\left(x,\sigma_x\left(1\right)\right) \leq \cdots \leq \Delta\left(x,\sigma_x\left(S\right)\right)~.
\end{align}
If $s$ is the closest activated slot to $x$, then all the slots closer to $x$ must be deactivated. This observation leads us to rewrite $\cL_{\XfitsY}$ as follows:
\begin{align}
    \!\!
    \cL_{\XfitsY}(\cM, \bX)\!=\!
    \frac{1}{|\bX|}\!
    \sum_{x\in\bX}
    {\sum_{s=1}^S}
    \Delta(x, s)
    \alpha_{s}\!\!\!\!\!\!
    \prod_{r<\sigma_x(s)}\!\!
    \!\!\!\!(1-\alpha_{\sigma^{-1}_x(r)})
    ~,
    \!\!
    \label{eq:XfitsY}
\end{align}
with $\sigma^{-1}$ the inverse permutation of $\sigma$.

\begin{figure*}[t]
    \centering
    \newcommand{\addci}[1]{\includegraphics[width=.165\textwidth]{images/lidarhd_compare/#1.jpg}}
\setlength{\tabcolsep}{1pt}
\begin{tabular}{@{}cccccc@{}}
    Input RGB & Input Intensity & \textbf{Ours}  & \DTI \cite{monnier2021unsupervised} & AtlasNet~v2 \cite{deprelle2019learning} & SuperQuadrics \cite{paschalidou2019superquadrics}\\
    \addci{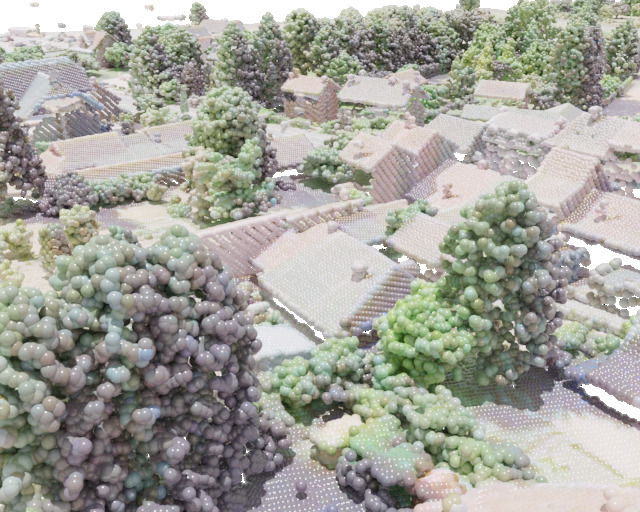} &
    \addci{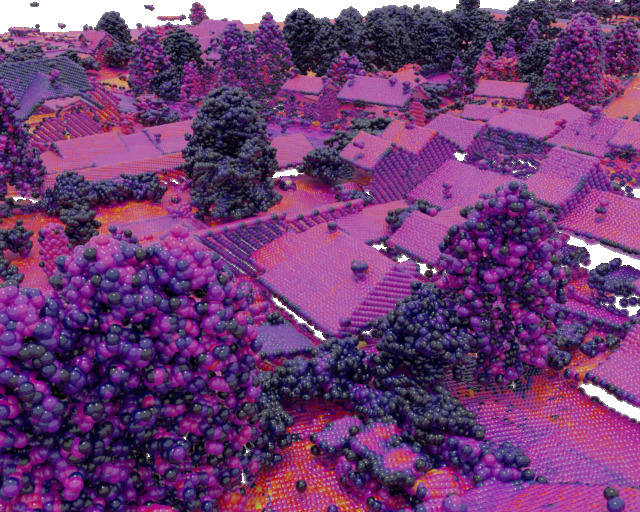} &
    \addci{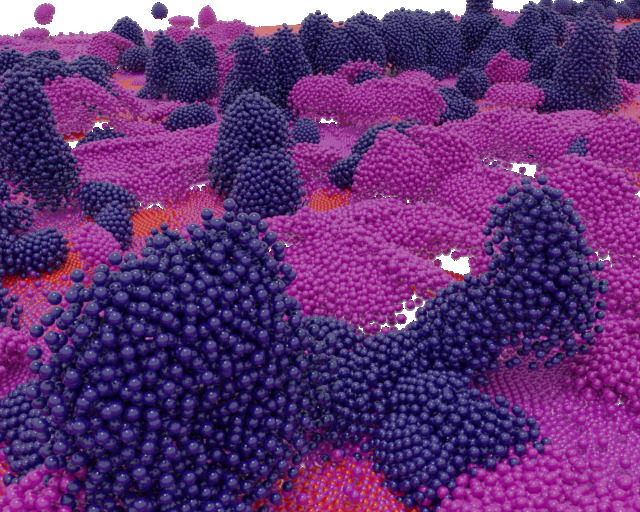} &  
    \addci{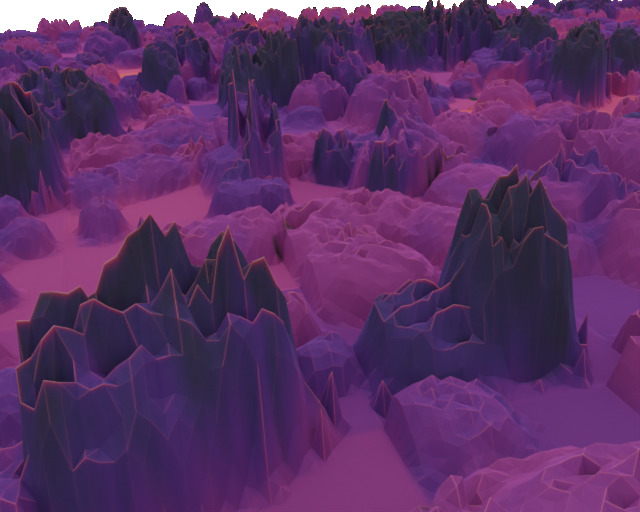} & 
    \addci{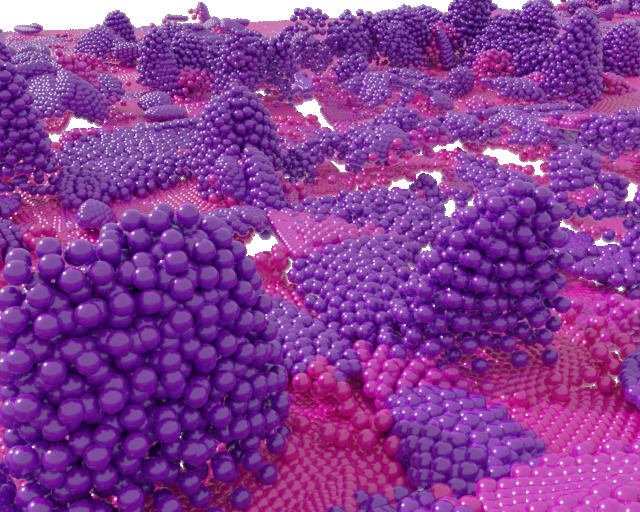} &
    \addci{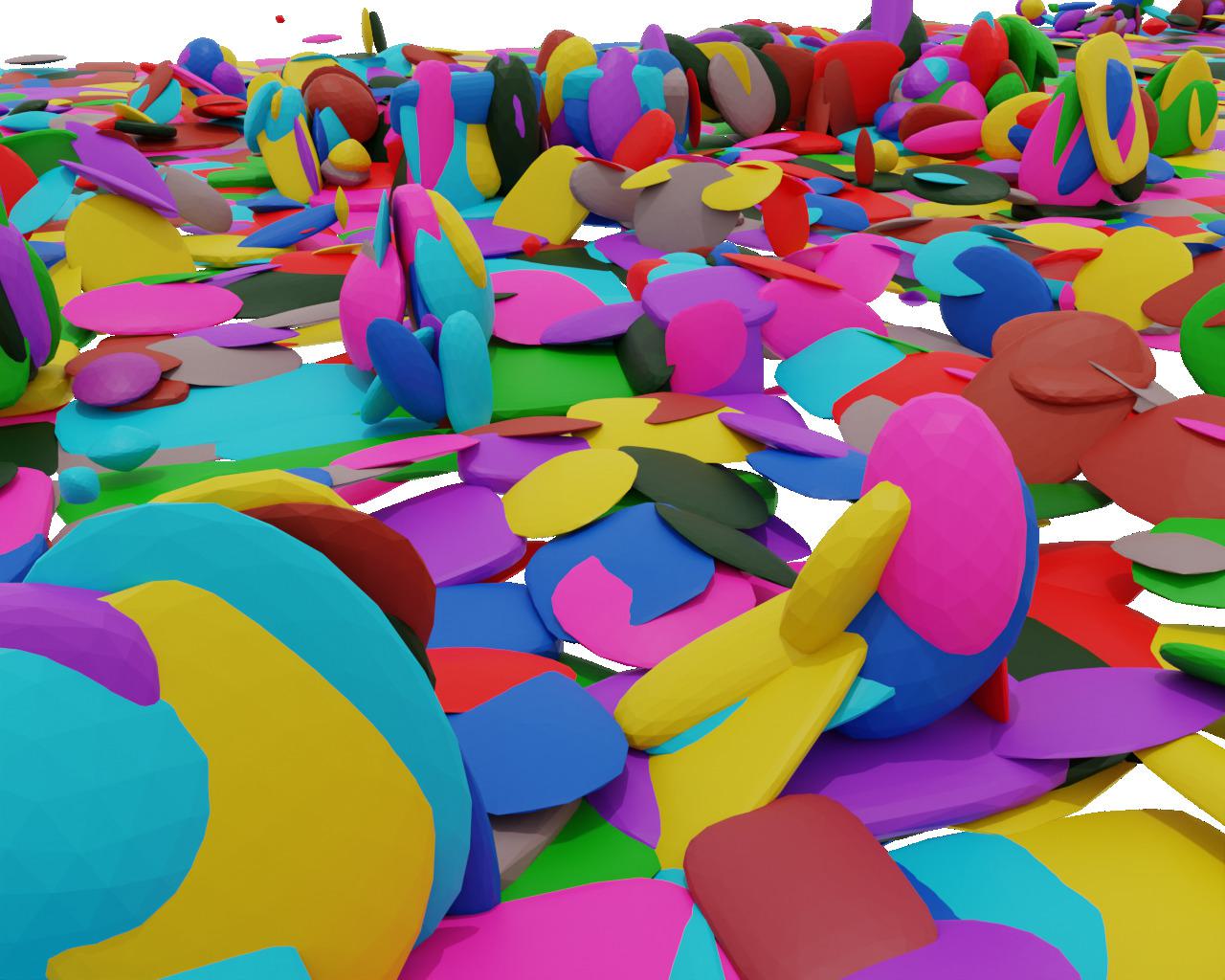}\\
    \addci{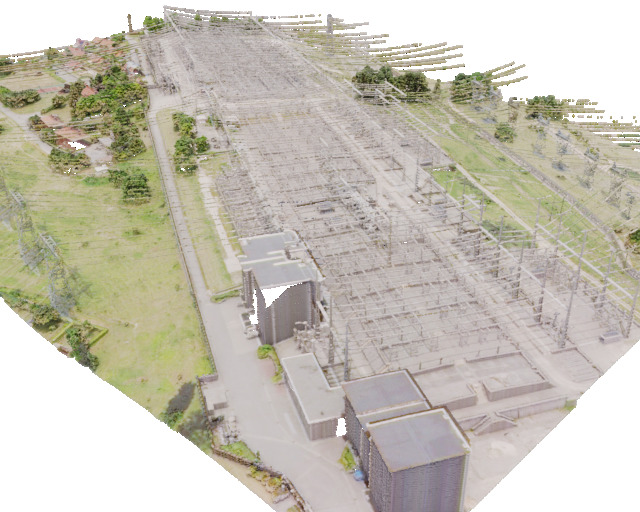}&
    \addci{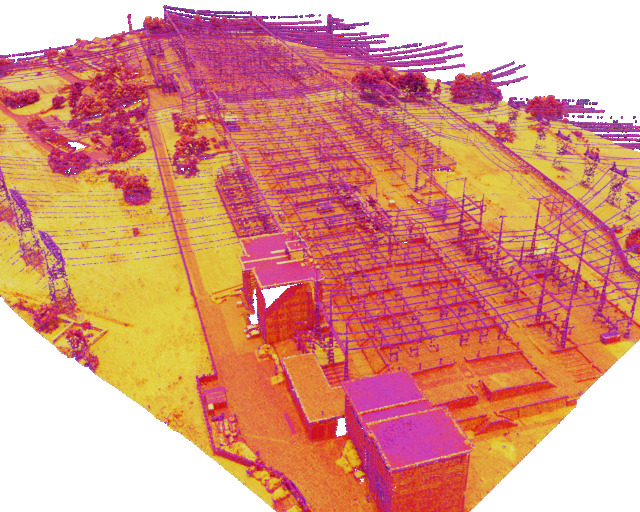}&
    \addci{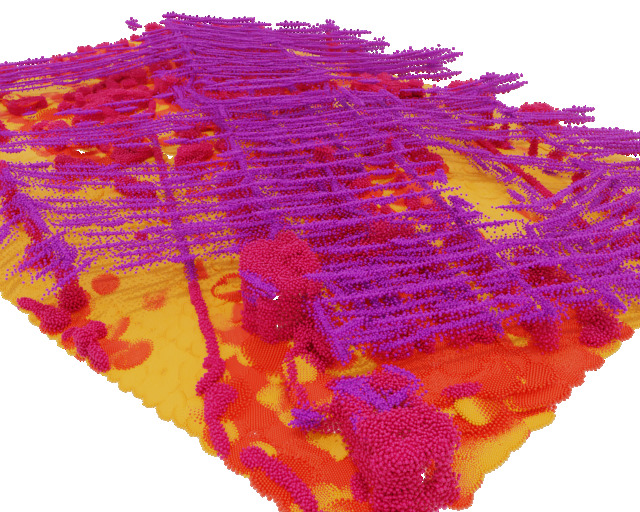} & 
    \addci{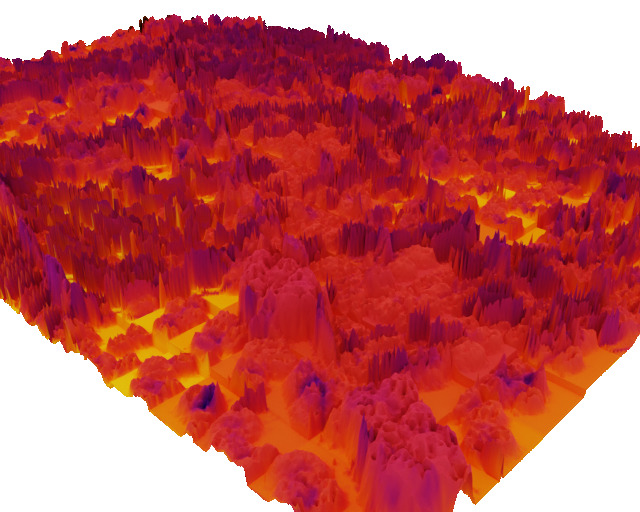} & 
    \addci{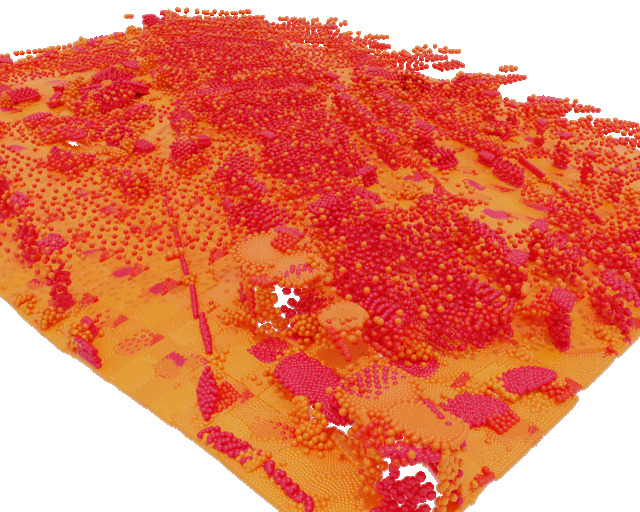} &
    \addci{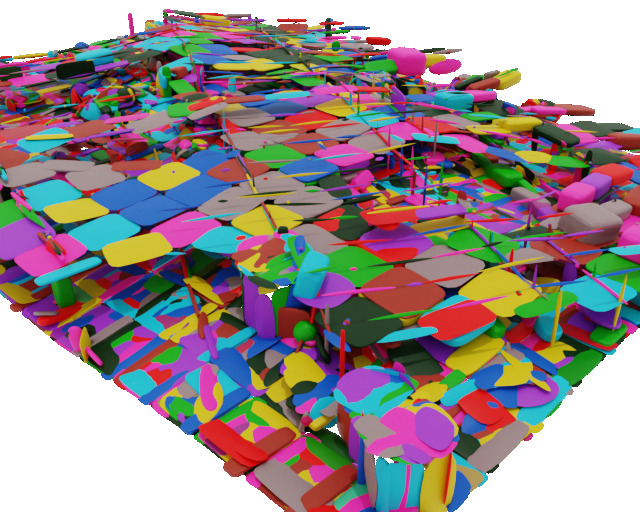}\\
\end{tabular}\vspace{-.75em}
    \caption{\textbf{Reconstruction Quality.}~We show two partial scenes with their RGB and  intensity values, as well as their reconstruction by our method and competing models.  We use the prototypes' intensity to color the points or pixels. As SuperQuadrics does not model the intensity, we use a random colour for each quadric.}
    \label{fig:comparison}
    \vspace{\vspacevalue em}
\end{figure*}

\paragraph{Regularization losses.}{As is often necessary with fully unsupervised reconstruction methods \cite{monnier2020dticlustering,loiseau2021representing,monnier2021unsupervised}, we define several regularization losses implementing priors on the model output to prevent degenerate local minima:
\begin{itemize}[itemsep=0em, wide, labelwidth=!, labelindent=0pt, topsep=0em]
    \item To encourage the slot activation to be sparse, we use the following loss penalizing slot activation:
    \begin{align}
    \cL_\text{act}(\cM)=\sum_{s=1}^S {\mathbb{E}_\bX\left[\alpha_s\right]}~.
    \end{align}
    \item To avoid slots and prototypes that are never used, we use the following losses, which we compute batch-wise:
    \begin{align}
        \cL_\text{slot}(\cM)
        &=
        -\sum_{s=1}^S\min
        \left(
        {\frac{\mathbb{E}_\bX\left[{\alpha_s}\right]}{\sum_{t=1}^S\mathbb{E}_\bX\left[{\alpha_t}\right]}}
        ,{\epsilon_S}
        \right)~,\\
        \cL_\text{proto}(\cM)&=
        -
        \sum_{k=1}^K
        \min\left(
            {\frac{ \mathbb{E}_\bX\left[{\sum_{s=1}^S\beta_s^k}\right]
            }{ {\sum_{s=1}^S\mathbb{E}_\bX\left[\alpha_s\right]}
            }},\epsilon_K
        \right)~,
    \end{align}
    with $\epsilon_S$ and $\epsilon_K$ hyperparameters setting the smallest acceptable relative use frequency for a slot or prototype.
\end{itemize}
}
The full regularization loss is a weighted sum of the three losses described above:
\begin{align}
    \cL_\text{reg} = \lambda_\text{act}\cL_\text{act} + \lambda_\text{slot}\cL_\text{slot} + \lambda_\text{proto}\cL_\text{proto}~,
\end{align}
where we use $\lambda_\text{act}=10^{-4}$, $\lambda_\text{slot}=\lambda_\text{proto}=0.1$ and $\epsilon_S=\epsilon_K=0.1$ in all our experiments on the \DATASETNAME. 

\subsection{Training and Implementation Details}

\paragraph{Model configuration.}We process the input point cloud $\bX$ using a similar architecture as in~\cite{paschalidou2019superquadrics}. We first voxelize it with a $64\times64\times64$ grid and map it to a vector using a sequence of $6$ 3D sparse convolutions~\cite{spconv2022} and $6$ strided convolutions.  The resulting representation is then transformed using one linear layer for each slot. These features are decoded by simple 2-layer MLPs: one generates the distribution parameters $\alpha_s$ and $\beta_s^k$, and the other ones the parameters of the 3D transformations $\cT_s(X)$. The transformations include an anisotropic scaling, a $y$-axis tilt of $\pm\sfrac{\pi}{10}$, a rotation around the $z$-axis, and a translation, in this specific order. We use $S=64$ slots and $K=6$ prototypes as default parameters. See the supplementary material for details.

The intensity of the return signal of each point is available in the LiDAR scans. We associate each prototype with a single learnable intensity parameter and perform the Chamfer distance (\eqref{eq:chamfer}) in 4 dimensions: spatial coordinates normalized to $[0,1]^3$ and intensity to $[0,0.1]$.

\paragraph{Curriculum learning.}The network predicts simultaneously the slot's probability distributions and their deformations. This results in many concurrent degrees of freedom and can make the training process unstable. Therefore, following Monnier~\etal~\cite{monnier2021unsupervised} and Loiseau~\etal~\cite{loiseau2021representing}, we implement a multi-stage curriculum learning strategy. We first initialize the prototypes as point clouds uniformly sampled from a random cuboid and gradually unfreeze the model parameters in the following order: (i) translation, rotation, tilt, slot activation, and choice of prototype; (ii) intensities of the prototypes, when available; (iii) scales of the prototypes; (iv) positions of the prototypes' 3D points; (v) anisotropic scalings of the prototypes. As shown in \secref{sec:results}, each step of this curriculum scheme improves the performances.

\begin{table*}[t]
    \caption{\textbf{Results on the \DATASETNAME.}~We report the quality of the reconstruction (Cham.) and semantic segmentation (mIoU) on each of the scenes of our \DATASETNAME. While our method does not always provide the most faithful reconstructions, it leads to the most accurate point classification. }
    \centering
    \resizebox{\textwidth}{!}{
\newcommand{\PreserveBackslash}[1]{\let\temp=\\#1\let\\=\temp}
\newcolumntype{C}[1]{>{\PreserveBackslash\centering}p{#1}}
\begin{tabular}{@{}lC{.9cm}C{.75cm}c@{~~}cc@{~~}cc@{~~}cc@{~~}cc@{~~}cc@{~~}cc@{}c}\toprule
& \multirow{2}{*}{\rotatebox{45}{{Rec.}}} & \multirow{2}{*}{\rotatebox{45}{{Semantic}}} &\multicolumn{2}{c}{\cropfield} &\multicolumn{2}{c}{\forest}&\multicolumn{2}{c}{\greenhouse} 		&\multicolumn{2}{c}{\boats}		&\multicolumn{2}{c}{\powerplant}	&\multicolumn{2}{c}{\nimes}&\multicolumn{2}{c}{\windturbine}	\\\cmidrule(lr){4-5}\cmidrule(lr){6-7}\cmidrule(lr){8-9}\cmidrule(lr){10-11}\cmidrule(lr){12-13}\cmidrule(lr){14-15}\cmidrule(lr){16-17}
& & &Cham.	&mIoU	&Cham.	&mIoU	&Cham.	&mIoU	&Cham.	&mIoU	&Cham.	&mIoU	&Cham.	&mIoU   &Cham. &mIoU	\\\midrule
k-means (i,z)~\cite{kmeans} & \xmark & \cmark	                                &---	&93.8	&---	&71.5	&---	&39.3	&---	&41.4	&---	&42.8	&---	&56.5	&---  & 87.6  \\
SuperQuadrics~\cite{paschalidou2019superquadrics} & 3D & \xmark & 0.86	&---	&1.04	&---	&0.60	&---	&0.93	&---	&0.58   &---	&0.40	&---	&13.5 & ---	\\
\DTI~\cite{monnier2021unsupervised} & 2.5D+i  & \cmark         &6.10	&83.2	&14.59	&40.2	&5.36	&42.0	&6.16	&41.4	&5.36	&29.0	&2.99	&47.3	&36.19	& 25.9\\
AtlasNet~v2~\cite{deprelle2019learning}  & 3D+i & \cmark  &1.07	&43.1   &1.58	&71.4	&0.56	&49.1   &\bf0.73	&42.1	&0.45   &41.6	&0.63	&48.8   &9.47 & 48.1\\
\bf Ours
& 3D+i & \cmark     & \bf 0.72 & \bf 96.9 & \bf 0.88 & \bf83.7 & \bf0.40 & \bf 91.3 & 0.82 & \bf 78.7 & \bf0.44 & \bf52.2 & \bf0.29 & \bf83.2 & \bf6.65 & \bf 93.4\\
\bottomrule
\end{tabular}
}
    \label{tab:multidti3d:lidarhd:results}
    \vspace{\vspacevalue em}
\end{table*}

\begin{table}[t]
    \caption{\textbf{\DATASETNAME.}~Our proposed dataset is composed of 7 diverse scenes acquired by aerial LiDAR.
    }
    \centering
    \renewcommand{\arraystretch}{1.}
\begin{tabularx}{\linewidth}{@{}lrrrc@{}}\toprule
    \multirow{2}{*}{Name}            & Surface  & \# points     & annotation &   num. of          \\
    & in km$^2$ & $\times10^6$ & ratio in \% & classes \\\midrule
    \cropfield	    & $1.1$	            & $19.7$	    & $77.4$	& 2                   \\
    \forest & $1.1$	            & $46.7$	    & $97.8$	& 2                       \\
    \greenhouse	    & $0.1$	            & $1.3$	    & $95.6$	& 3\\
    \boats	        & $0.1$	            & $0.5$       & $92.7$    & 2                     \\
    \powerplant	    & $0.2$	            & $8.6$	    & $78.4$	& 4                   \\
    \nimes	        & $1.1$	            & $15.7$	    & $95.9$	& 3                   \\
    \windturbine	    & $4.2$	            & $5.6$	    & $99.8$  & 3                      \\\midrule
    Total	&$7.7$	& $98.3$	& $91.6$	& ---\\
    \bottomrule
\end{tabularx}

    \label{tab:lidarhd:scenes}
    \vspace{\vspacevalue em}
\end{table}

\paragraph{Prototypes selection.} We automatically select the number of prototypes for a complete scene using a simple greedy algorithm. We measure the increase of reconstruction loss when preventing the model from selecting each prototype individually. We remove the prototype with the lowest increase if it is lower than $5$\%, and iterate.

\paragraph{Implementation details.}Our model is trained separately for each scene by randomly sampling square patches. During training, the patches are subsampled to a maximum $10^5$ points. Each stage of the curriculum is trained until convergence. We use the ADAM optimizer~\cite{kingma2014adam} with a learning rate of $10^{-4}$ and default parameters. See the supplementary material for preprocessing, training and evaluation details.
\section{Results}\label{sec:results}

In this section, we assess the ability of our method to parse complex 3D aerial data. We give in \secref{sec:dataset} an overview of our proposed dataset of aerial LiDAR scans. In \secref{sec:baselines}, we then discuss our evaluation metrics and baselines. Finally, we present a quantitative (\secref{sec:quant}) and qualitative (\secref{sec:quali}) analysis of our results. 

\subsection{\DATASETNAME}
\label{sec:dataset}
We introduce a new dataset to train and evaluate parsing methods on large, uncurated aerial LiDAR scans. We use data from the French Mapping Agency associated to the LiDAR-HD project \cite{lidarhd}. Each scan is composed of several airborne LiDAR acquisitions taken at different angles, leading to a minimum resolution of $20\ \text{points}/\text{m}^2$. The points are associated with their laser reflectance (intensity), and colorized based on asynchronous aerial photography. 

We selected $7$ scenes, covering over $7.7$km$^2$ and a total of $98$ million 3D points, with diverse content and complexity, such as dense habitations, forests, or complex industrial facilities. We associate most 3D points with a coarse semantic label, such as ground, building, or vegetation. The characteristics of the scenes are detailed in Table~\ref{tab:lidarhd:scenes} and each is visualized in \figref{fig:lidarhd:scenes}.
\subsection{Evaluation Metrics and Baselines}
\label{sec:baselines}

We quantitatively evaluate the performances for reconstruction and semantic segmentation of our model and several unsupervised scene decomposition approaches. 

\paragraph{Evaluation metrics.} As our goal is to summarize a point cloud using few prototypes, the quality of the reconstruction is critical. We measure it with the symmetric Chamfer distance (``Cham." in the Tables) between the input and the output point clouds of our model.

By associating a class to each prototype's points, we can propagate labels from the reconstruction to the input cloud and perform semantic segmentation. We evaluate the quality of this segmentation with the {class-averaged} Intersection-over-Union (mIoU) metric.

In a practical scenario, an operator can manually annotate the points of the 3D prototypes, allowing for the segmentation of the entirety of $\bX$ with minimal effort. To perform automatic evaluation, we follow the standard practice for evaluating clustering and unsupervised segmentation methods~\cite{ji2019invariant,kosiorek2019stacked,monnier2020dticlustering}: we assign to each prototype's point the most frequent class of its closest point of the input after the reconstruction.

\paragraph{Baselines.}
We adapt several unsupervised approaches for scene reconstruction and/or semantic segmentation tasks to provide comparisons for our approach:
\begin{itemize}[itemsep=0em, wide, labelwidth=!, labelindent=0pt, topsep=0em, parsep=0em]
    \item \textbf{k-means.}~We cluster the points of the input with the k-means algorithm~\cite{kmeans} using as many clusters as we use prototypes. We obtain the best results by using a combination of the point's intensity and elevation as features for clustering. We then assign to each centroid its most frequent class, and propagate this label to the entire cluster, leading to a semantic segmentation. This method does not reconstruct the input, but gives us a simple and surprisingly strong baseline for semantic segmentation.
    \item \textbf{SuperQuadrics revisited.}~We use the method of Paschalidou~\etal~\cite{paschalidou2019superquadrics} to learn to approximate scenes with an adaptive number of superquadrics \cite{barr1981superquadrics}. It provides a baseline for reconstruction and a qualitative comparison for instance segmentation, shown in supplementary material.
    \item \textbf{DTI-Sprites.}~We use the point cloud to construct a digital elevation model, \ie a 2.5D image of resolution $32\times32$ where each pixel has an elevation and intensity value. We adapt the unsupervised image decomposition approach of Monnier~\etal~\cite{monnier2021unsupervised} to break down this image into a set of 2.5D \emph{sprites}. We evaluate the reconstruction and segmentation by sampling $25$ 3D point per pixel, transferring the pixel's label to the points, and interpolating  their elevations.
    \item \textbf{AtlasNet v2.}~This extension \cite{deprelle2019learning} of AtlasNet \cite{groueix2018papier} uses a fixed number of learnable prototype point clouds to reconstruct its input. It can be evaluated for both reconstruction and semantic segmentation in a way similar to ours. We extend it to handle intensity in a manner akin to our approach, which improves its segmentation results.
\end{itemize}
Similar to our method, we train all baselines except k-means by sampling square patches in each scene. \figref{fig:comparison} shows the output of the reconstruction methods.
\subsection{Quantitative Results}
\label{sec:quant}
We compare the performance of our approach with the proposed baselines on the Earth Parser Dataset, as well as two publicly available datasets.

\begin{table}[t]
    \caption{\textbf{Ablation Study.}~We evaluate the effect of our prototype selection post-processing, our model's degrees of freedom, and our different regularization losses. }
    \centering    
    \def\mycorner{
\scalebox{0.5}{\tikz \draw [->, thick] (0,.5) -- (0,0) -- (.5,0);}
}

\renewcommand{\DTstyle}{\textrm}

\begin{tabular}{@{}l@{}lc@{~~}cc@{~~}c@{}}\toprule
       & &\multicolumn{2}{c}{\nimes}&\multicolumn{2}{c}{\boats}\\\cmidrule(lr){3-4}\cmidrule(lr){5-6}
       & & Cham. & mIoU & Cham. & mIoU \\\midrule

& {\METHODNAME}
& 0.29 & 83.2 & 0.82 & \bf78.7\\
& \hspace{0.1em}\mycorner w/o {post-processing}  & 0.28 & \bf83.7 & 0.96 & 78.3\\\cdashline {1-6} 
\multirow{4}{*}{\rotatebox{90}{
\small{expressivity}}}
& \hspace{0.4em}\mycorner w/o aniso-scale  & 0.33 & 82.4 & 1.04 & 67.2\\
& \hspace{0.6em}\mycorner w/o prototypes  & 0.36 & 68.3 & 1.07 & 42.8\\
& \hspace{0.8em}\mycorner w/o scales  & 0.55 & 58.9 & 1.33 & 40.8\\
& \hspace{1em}\mycorner w/o intensities & 0.55 & 58.7 & 1.09 & 40.8\\\cdashline{1-6} 

\multirow{3}{*}{\rotatebox{90}{\!\small{losses}}}
& \hspace{0.4em}\mycorner w/o $\cL_\text{act}$ & \bf0.17 & 54.1 & \bf0.80 & 56.9\\
& \hspace{0.4em}\mycorner w/o $\cL_\text{slot}$ & 0.25 & 77.8 & 0.81 & 43.7\\
& \hspace{0.4em}\mycorner w/o $\cL_\text{proto}$ & 0.28 & 57.2 & 0.97 & 40.7\\
\bottomrule\end{tabular}
    \label{tab:multidti3d:lidarhd:ablation}
\end{table}

\paragraph{\DATASETNAME.} We provide quantitative reconstruction and semantic segmentation results in \tabref{tab:multidti3d:lidarhd:results}, and illustrations in \figref{fig:comparison}. Despite being highly constrained, our model yields the best reconstruction in $6$ out of $7$ scenes. Moreover, we significantly outperform the other methods for semantic segmentation across all scenes. 

Despite its simplicity, the k-means baseline provides strong semantic segmentation performance, beating the other baselines in $5$ of $7$ annotated scenes. DTI-Sprites~\cite{monnier2021unsupervised} has lower reconstruction and segmentation quality, which is expected as it models a 3D point cloud in 2.5D. AtlasNet~v2~\cite{deprelle2019learning} provides good reconstructions but segmentation fails for scenes such as Crop Fields or Urban due its inability to adapt its prototype usage to the input. On the contrary, SuperQuadrics \cite{paschalidou2019superquadrics} can adjust the number of superquadric it uses and, to some degree, their shape. However, this method uses a single parametric family for all prototypes and fails to reconstruct complex real-world scenes such as Power Plant.
Thanks to its probabilistic slot selection, our method can handle inputs with a varying number of objects using only a small set of learned prototypical shapes.

\begin{table}[t]
    \centering
    \caption{\textbf{Synthetic Shapes.}~We train our method on all planes from ShapeNet-Part~\cite{savva2015semgeo}, with random rotations around $z$-axis. We show the reconstruction of an input plane and the prototypes learned on the dataset.}
    \newcommand{\planerec}[2]{\includegraphics[width=.17\linewidth]{images/shapenetsem/airplanes_v2/#1/snsem_#2_reconstruction_point_y_0.jpg}}

\newcommand{\planeproto}[2]{\includegraphics[width=.17\linewidth]{images/shapenetsem/airplanes_v2/#1/snsem_#2_prototypes_point_y_0.jpg}}

\newcommand{\planeleg}[1]{\rotatebox{90}{\small{#1}}}

\newlength{\SNraise}
\setlength{\SNraise}{.4cm}
 
{\small{\begin{tabular}{@{}lccc@{}}\toprule
    & AtlasNet~v2~\cite{deprelle2019learning} & SuperQuadrics~\cite{paschalidou2019superquadrics} & \textbf{Ours} \\\midrule
    \raisebox{\SNraise}{Recons.} & \planerec{rotated}{anv2} & \planerec{rotated}{sq} & \planerec{rotated}{ours} \\
    \raisebox{\SNraise}{Protos.} & \raisebox{0.2cm}{\planeproto{rotated}{anv2}} & \raisebox{\SNraise}{---} & \planeproto{rotated}{ours} \\
    \rotatebox{0}{Cham.} & 1.46 & 2.91 & \bf 1.34\\
    \rotatebox{0}{mIoU} & 34.5 & --- & \bf 68.6 \\
    \bottomrule
\end{tabular}}}
    \label{tab:shapenetsem}
    \vspace{\vspacevalue em}
\end{table}

\begin{figure}%
    \centering
  \begin{tabular}{@{}c@{}c@{}} 
    \raisebox{1.25cm}{\begin{subfigure}[t]{0.56\linewidth}
        {\small{
        \begin{tabular}{@{}l@{~~}c@{~~}c@{}}\toprule
        & Cham. & mIoU \\\midrule
        k-means (i,z)~\cite{kmeans} & --- & 52.7 \\
        Superquadrics~\cite{paschalidou2019superquadrics} &  0.33 & --- \\
        DTI-Sprites~\cite{monnier2021unsupervised} & 2.25 & 53.6 \\
        AtlasNet~v2~\cite{deprelle2019learning} & 0.31 & 54.9\\
        \textbf{Ours} & \textbf{0.29} & \textbf{82.7} \\\bottomrule
    \end{tabular}}}
    \end{subfigure}}
    &
    \begin{subfigure}[t]{0.45\linewidth}
        \includegraphics[width=\linewidth]{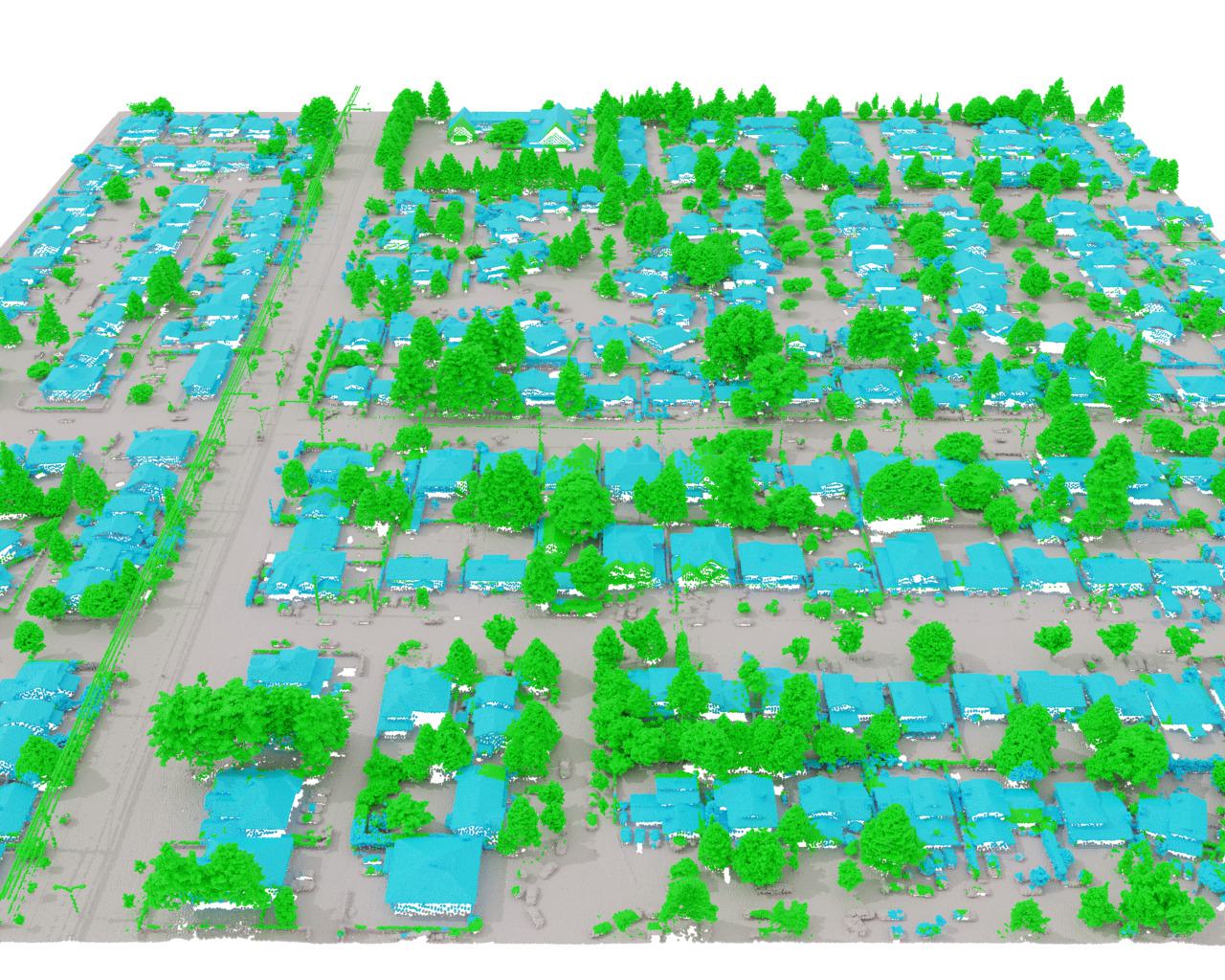}
    \end{subfigure}
\end{tabular}
    \caption{\textbf{Results on DALES~\cite{varney2020dales}.}~We report quantitative and qualitative results for one tile from DALES.}  
    \label{fig:resultDales}
    \vspace{\vspacevalue em}
\end{figure}

\paragraph{Ablation study.}
We evaluate the impact of various components of our model and report the results in \tabref{tab:multidti3d:lidarhd:ablation}. First, we observe that the prototype selection post-processing has limited impact on the quality of the prediction and reconstructions, but allows us to adapt and significantly decrease the number of prototypes used for each scene. Second, we evaluate the impact of reducing the expressivity of our model. We successively remove: (i) the anisotropic scaling of $\cT$, and the possibility of learning the prototype's (ii) points position, (iii) scale, and (iv) intensity. As expected, each degree if liberty removed decreases the quality of the 3D reconstruction and segmentation.

We study the impact of the different regularization introduced in \secref{sec:losses}. The losses related to slot activation have a marginal effect on reconstruction quality but significantly affect the performance of semantic segmentation. This suggests that using slots sparingly but equally is important for prototypes to specialize for specific objects, but is not necessary to the expressivity of the reconstruction model. 
The prototype activation loss prevents unused and possibly degenerate prototypes and therefore improves both segmentation and reconstruction.

We evaluate the impact of reducing the number of prototypes or slots. When we set $K=3$ instead of $6$, the average semantic segmentation mIoU drops by $26.3$ points,  while reconstruction is minimally affected, showing only a $2.3$\% increase in Chamfer distance. Conversely, setting $S=32$ instead of $64$ leads to a significant $+47.3\%$ increase of Chamfer distance, with a modest drop $0.4$ of semantic segmentation mIoU. These findings highlight the importance of maintaining sufficient diversity in the prototypes to enable specialization for different object types, whereas the number of slots strongly influences the expressiveness of the reconstruction model.

\paragraph{ShapeNet.}We  evaluate our model on $2690$ planes from ShapeNet-Part~\cite{savva2015semgeo}, whose points are annotated as \emph{wing}, \emph{engine}, \emph{tail}, or \emph{body}. We randomly rotate the shapes around the $z$-axis during training and evaluation. We report in~\tabref{tab:shapenetsem} the performances and reconstructions for our approach, AtlasNet~v2~\cite{deprelle2019learning}, and SuperQuadrics revisited~\cite{paschalidou2019superquadrics}. Our method handle rotations better than AtlasNet~v2, and manages to successfully locate the tail of the planes. While SuperQuadrics' reconstructions make sense qualitatively, they are worse in terms of accuracy than ours, and do not enable semantic segmentation.

\paragraph{DALES.}We also train and evaluate our model on DALES~\cite{varney2020dales}, a dataset of aerial LiDAR scans of urban area, with a restricted class set: \emph{ground}, \emph{vegetation}, and \emph{buildings}. Our model significantly outperform competing reconstruction-based approaches, and yields similar performance than for the {Urban} tile of the Earth Parser Dataset; see Figure~\ref{fig:resultDales}.
\subsection{Qualitative Results}
\label{sec:quali}

\begin{figure}
    \centering

\centering
\setlength{\tabcolsep}{3pt}
\begin{tabular}{@{}c@{~}c@{~}c@{}}
    \rotatebox{90}{~~~~~~~~~~~~\forest}
    &
    \includegraphics[width=.47\linewidth]{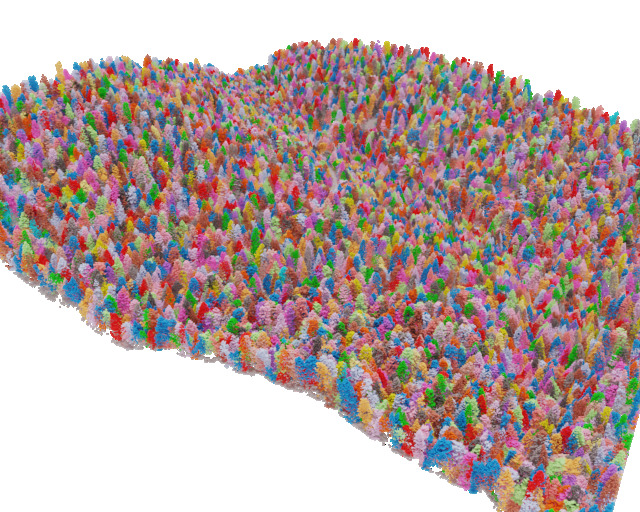}
    &
    \includegraphics[width=.47\linewidth]{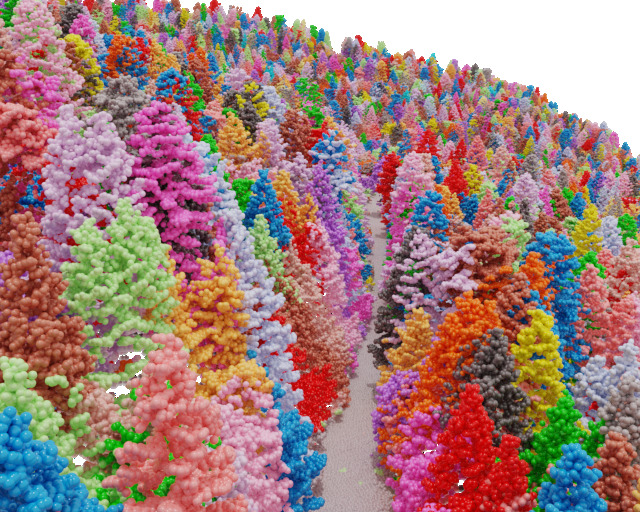}\\
    \rotatebox{90}{~~~~~~~~~~~~\boats}
    &
    \includegraphics[trim={0 0 0 0},clip,width=.47\linewidth]{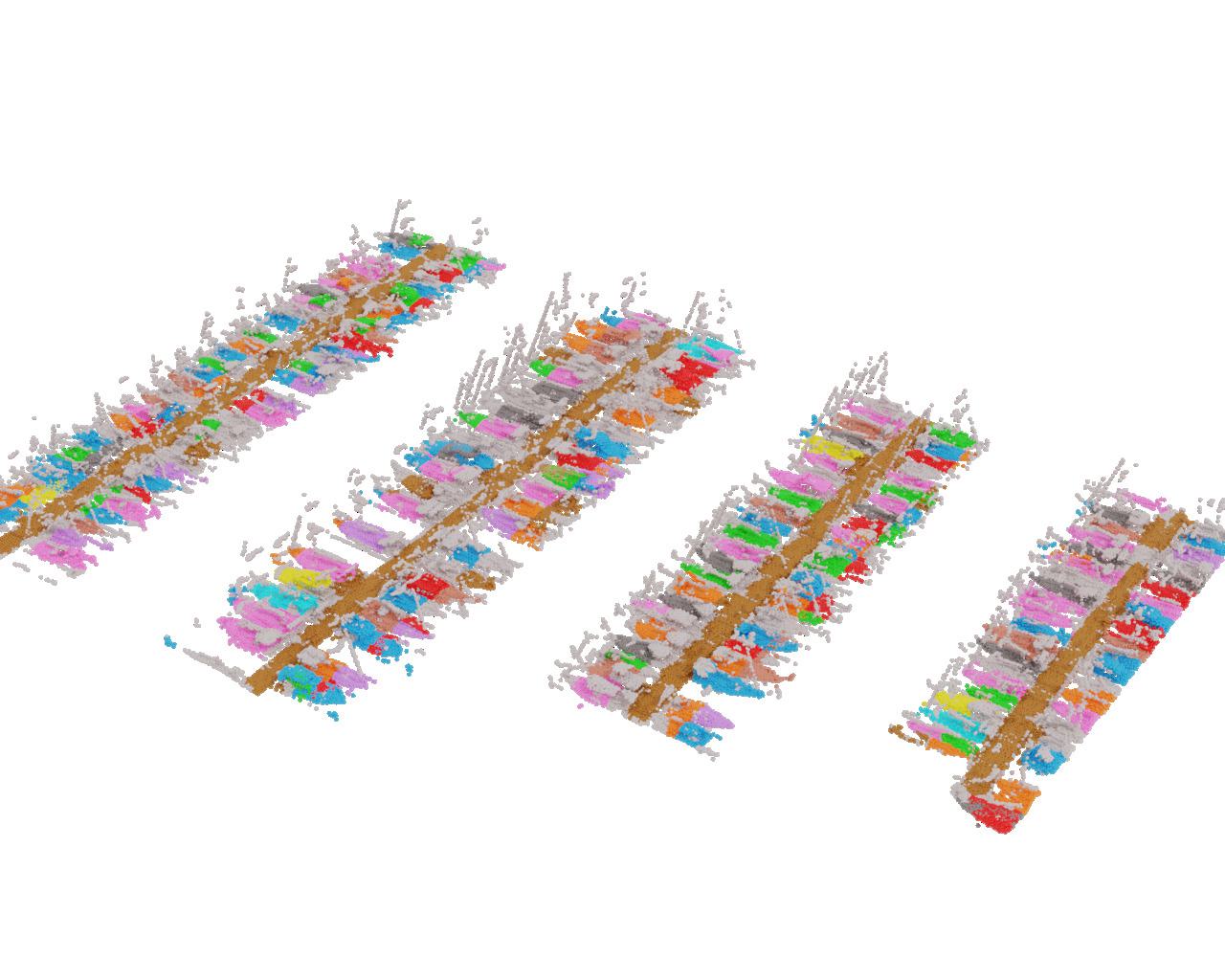}
    &
    \includegraphics[width=.47\linewidth]{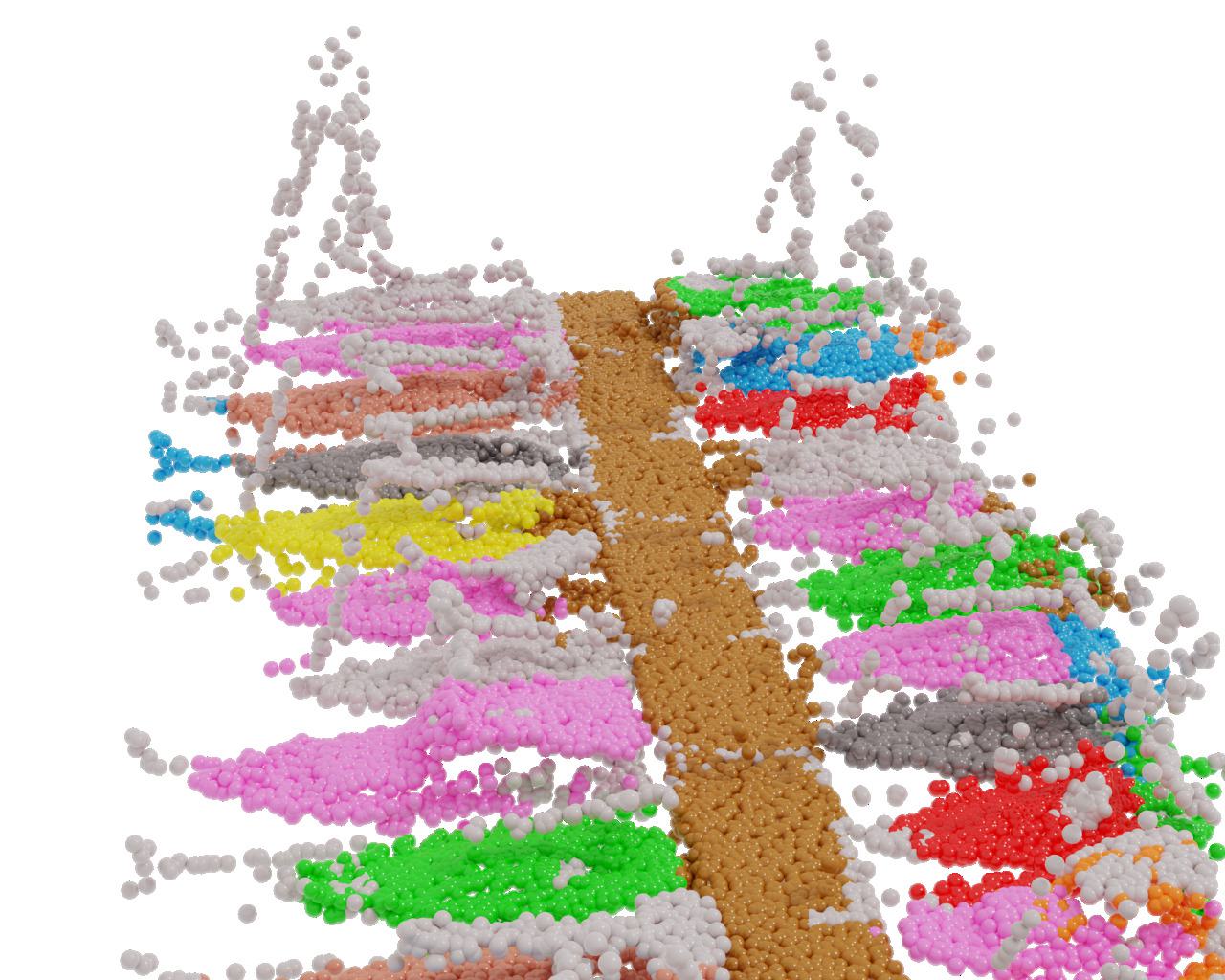}
\end{tabular}

    \caption{\textbf{Instance Segmentation.}~
    We can identify the reconstruction of each slot as a separate instance, allowing us to perform instance segmentation of complex data such as dense natural forests or a marina. For this visualization, we considered the points associated to ``trees" or ``boat hull" prototypes and color each of their instance randomly.
    }
    \label{fig:lidarhd:inst}
    \vspace{\vspacevalue em}
\end{figure}

\begin{figure}
    \centering

\newcommand{\pairProto}[1]{
     \includegraphics[width=.49\linewidth]{images/zooms/protos_sel_v3/#10000.jpg}
}


\renewcommand{\arraystretch}{1.}
\begin{tabular}{@{}c@{}c@{}}
     \pairProto{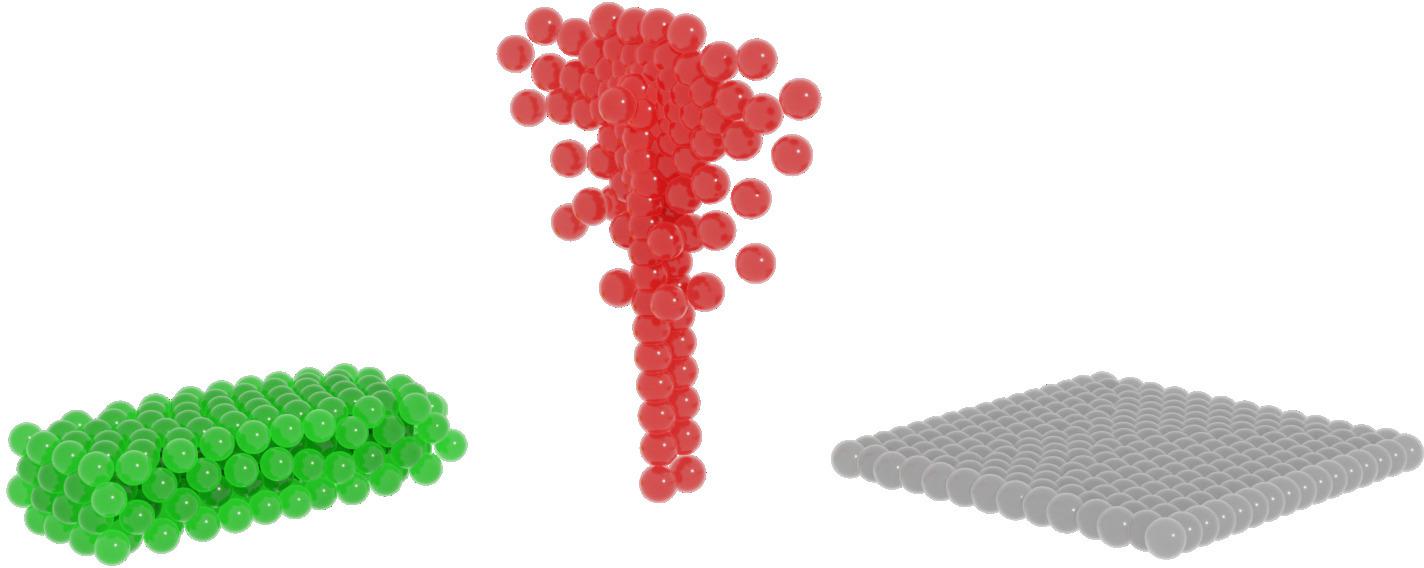} & \pairProto{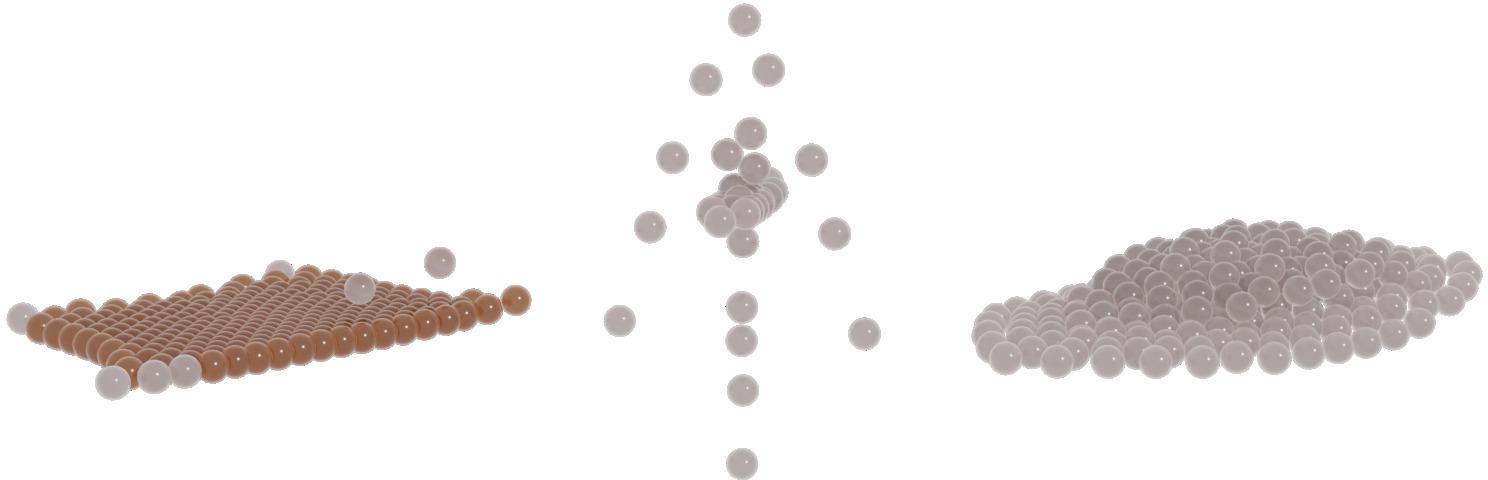}\\
     \windturbine~(3/5) & \boats~(3/3)\\
     \pairProto{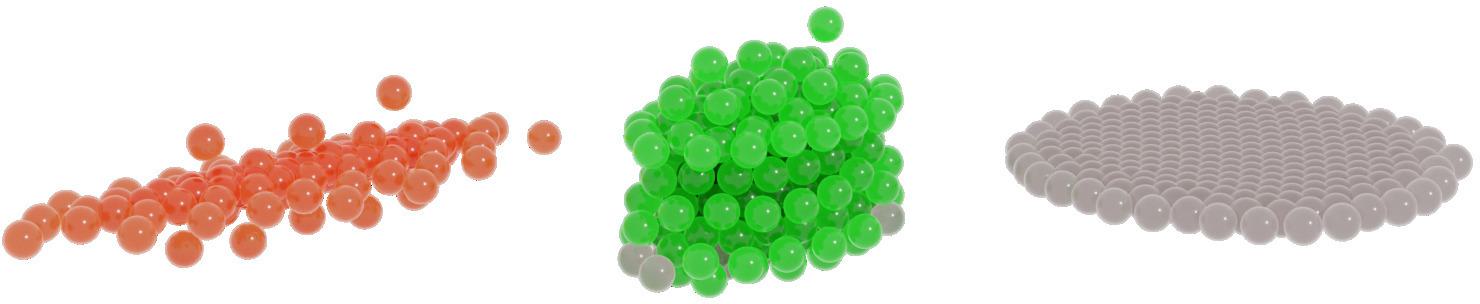}& \pairProto{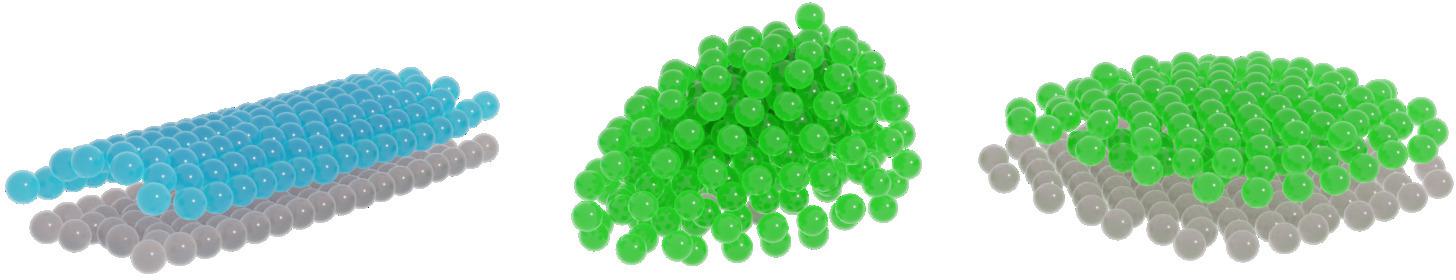}\\
     \powerplant~(3/4) & \greenhouse~(3/4)
\end{tabular}
    \caption{\textbf{Learned Prototypes.}~
We display three of the prototypes chosen during the selection process for various scenes.}
    \label{fig:lidarhd:protos}
    \vspace{\vspacevalue em}
\end{figure}

\paragraph{Instance segmentation.}We can perform instance segmentation simply by considering each slot as a different instance. This is particularly interesting for parsing natural woodlands, a key endeavor for forest management \cite{morsdorf2004lidar} and biomass estimation \cite{ferraz2016airborne}. While this task has a long history of handcrafted approaches \cite{vauhkonen2012comparative}, current deep learning approaches are mostly limited to artificial or low-density forests \cite{sun2022individual}. As shown qualitatively in \figref{fig:lidarhd:inst}, our \METHODNAME \ can learn without any supervision to separate individual trees in dense forests, or boats in a marina.
{We evaluate quantitatively the performance of our model in terms of instance segmentation, we counted manually $88$ boat instances in 10 distinct portions of the Marina scene. We then computed the Mean Relative Error (MRE) of the prediction given by the number of boat-like prototypes (prototype~\#3 in \figref{fig:lidarhd:protos}) in these zones. Our method yields a MRE of $7,4\%$, $2.6$  times lower than  DTI-Sprites~\cite{monnier2021unsupervised}, demonstrating its strong performances.}

\begin{figure}
    \centering

\newcommand{\pairGtSem}[2]{
    \rotatebox{90}{#1}
    &
    \includegraphics[trim={0 0 0 250px},clip,width=.47\linewidth]{images/lidarhd/inputs/point_y/#2.jpg}
    &
    \includegraphics[trim={0 0 0 250px},clip,width=.47\linewidth]{images/lidarhd/results/ours/point_y_pred/#2.jpg}
}

\centering
\setlength{\tabcolsep}{3pt}
\begin{tabular}{@{}c@{~}c@{~}c@{}}
    & Ground Truth & Prediction \\
    \pairGtSem{~~~~~~~~\nimes}{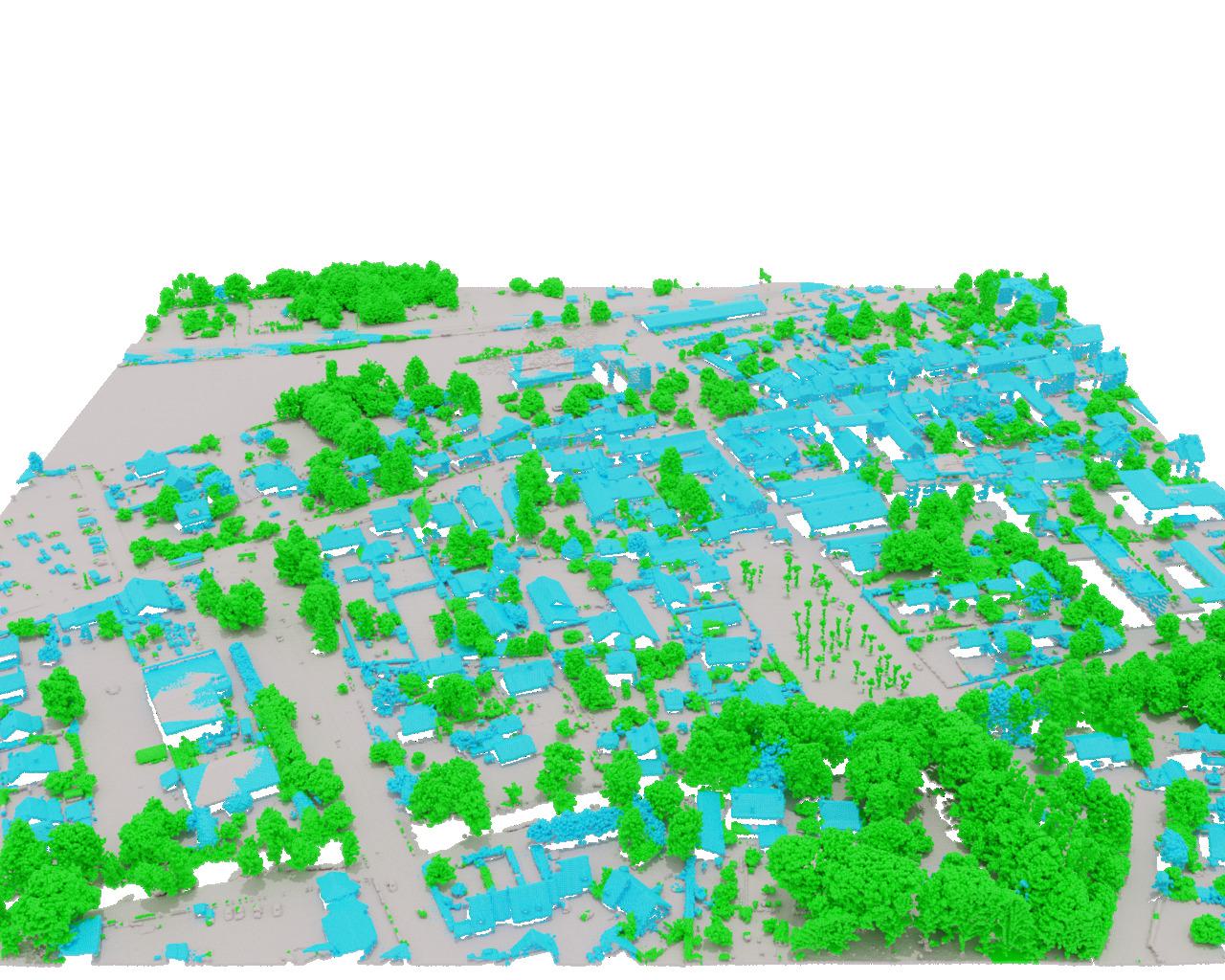}\\
    \rotatebox{90}{~~~~~~~~~\powerplant}
    &
    \includegraphics[width=.47\linewidth]{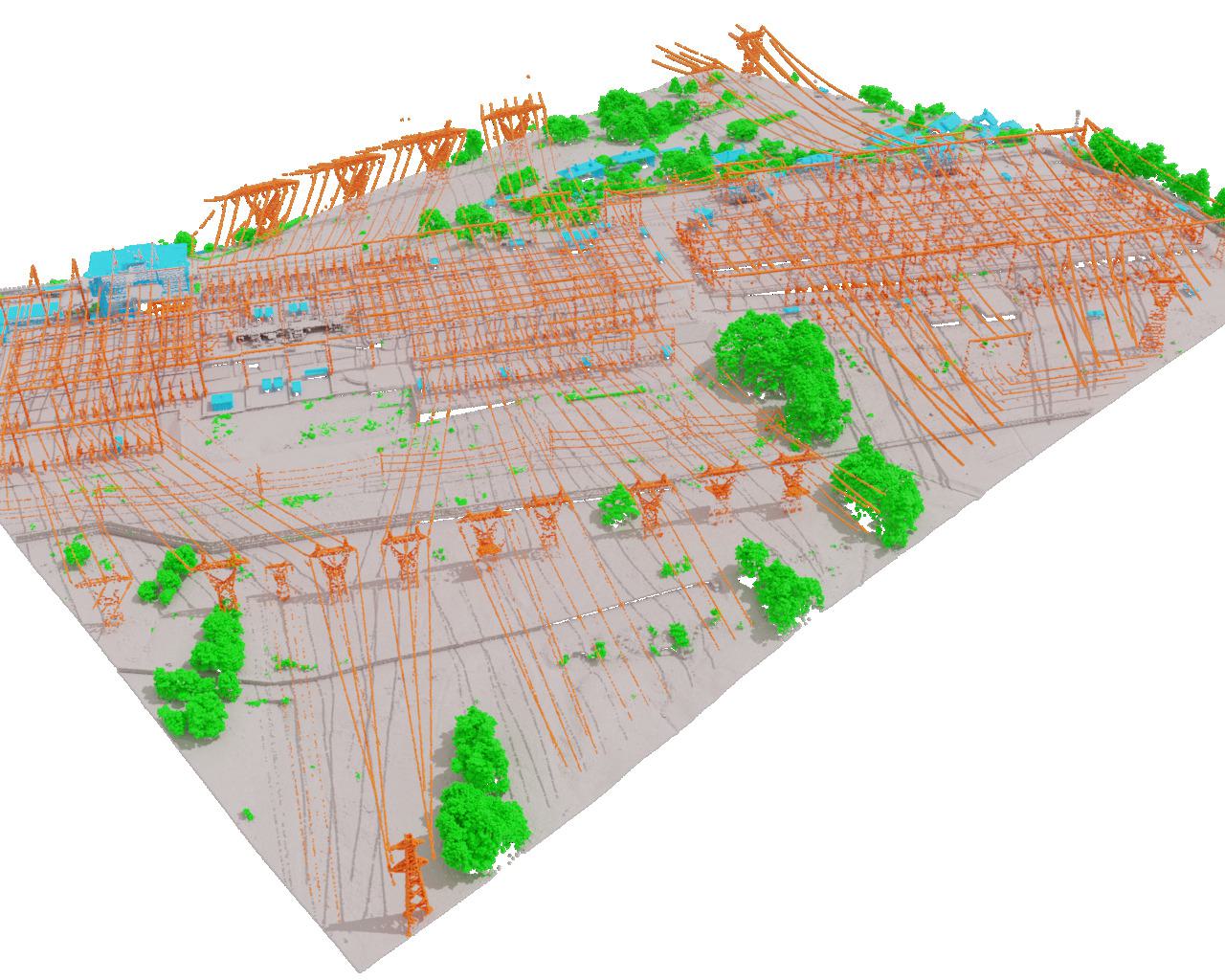}
    &
    \includegraphics[width=.47\linewidth]{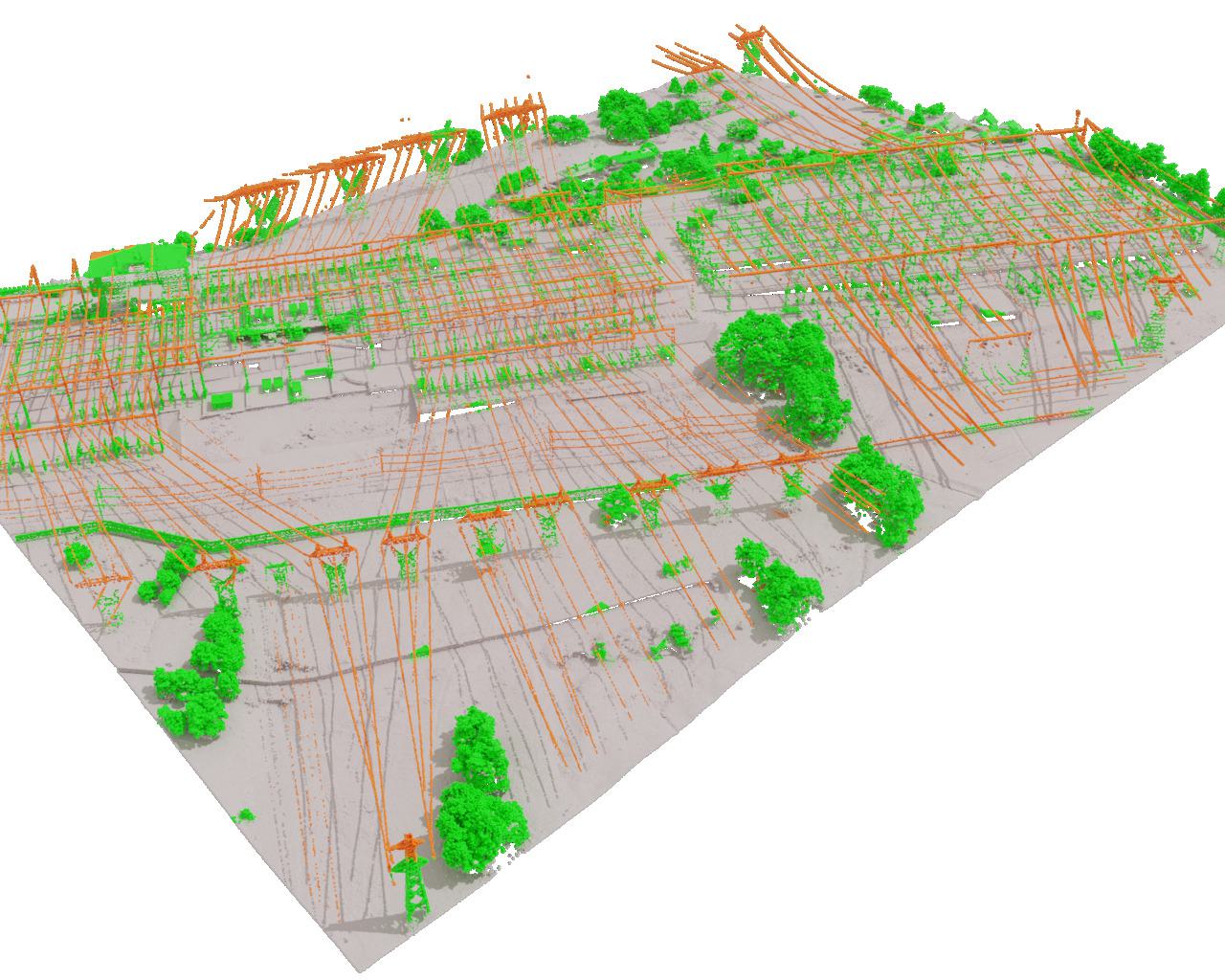}
\end{tabular}

    \caption{\textbf{Semantic Segmentation.}~Our model can perform semantic segmentation of large real-world scene based on annotated prototypes. Black points in the ground truth are unannotated.}
    \label{fig:lidarhd:semantic}    
    \vspace{\vspacevalue em}
\end{figure}

\paragraph{Semantic segmentation.}
After annotating the prototypes, our model can perform convincing semantic segmentation of complex scenes, as shown in \figref{fig:lidarhd:semantic}.

\paragraph{Interpretable prototypes.}In \figref{fig:lidarhd:protos}, we show prototypes learned on our \DATASETNAME \ with colors showing the associated semantic label for each point. These prototypes give at a glance insights on the content of these real-world scenes. Our model is able to learn a wide variety of shapes, such as boats' masts, wind turbines, or greenhouses. We also observe that the prototypes are typically associated with a single object type. Empirically, the average class distribution of the nearest neighbors of each prototypes’ point exhibits the same normalized entropy ($0.22$) as a Bernoulli distribution with probability $0.964$. The highly constrained nature of the deformations $\cT$ prevents slots from repurposing the same prototypes for different objects, making the learned prototypes easy to identify and manually annotate.

\section{Conclusion}\label{sec:conclusion}

We introduced a novel unsupervised method for parsing complex real-world aerial scans into simple parts using a small set of learned prototypical shapes. We demonstrate the quality and interpretability of our results on a novel dataset of aerial LiDAR scans. To the best of our knowledge, we are the first to demonstrate the possibility of performing deep unsupervised 3D shape analysis on such a challenging real-world dataset. We believe that our results open new perspectives for computer-assisted environment monitoring and economic intelligence.
\paragraph*{Acknowledgements}{This work was supported in part by ANR project READY3D ANR-19-CE23-0007, ANR under the France 2030 program under the reference ANR-23-PEIA-0008, and was granted access to the HPC resources of IDRIS under the allocation 2022-AD011012096R2 made by GENCI. The work of MA was partly supported by the European Research Council (ERC project DISCOVER, number 101076028). The scenes of \DATASETNAME \ were acquired and annotated by the LiDAR-HD project~\cite{lidarhd}. We thank Zenodo for hosting the dataset. We thank Zeynep Sonat Baltaci, Emile Blettery, Nicolas Dufour, Antoine Guédon, Helen Mair Rawsthorne, Tom Monnier, Damien Robert, Mathis Petrovich and Yannis Siglidis for inspiring discussions and valuable feedback.}

{\small
\bibliographystyle{templates/cvpr2024/ieeenat_fullname}
\bibliography{egbib}

\begin{thebibliography}{82}
\providecommand{\natexlab}[1]{#1}
\providecommand{\url}[1]{\texttt{#1}}
\expandafter\ifx\csname urlstyle\endcsname\relax
  \providecommand{\doi}[1]{doi: #1}\else
  \providecommand{\doi}{doi: \begingroup \urlstyle{rm}\Url}\fi

\bibitem[Alliegro et~al.(2021)Alliegro, Boscaini, and Tommasi]{alliegro2021joint}
Antonio Alliegro, Davide Boscaini, and Tatiana Tommasi.
\newblock Joint supervised and self-supervised learning for {3D} real world challenges.
\newblock In \emph{ICPR}, 2021.

\bibitem[Armeni et~al.(2016)Armeni, Sener, Zamir, Jiang, Brilakis, Fischer, and Savarese]{armeni20163d}
Iro Armeni, Ozan Sener, Amir~R Zamir, Helen Jiang, Ioannis Brilakis, Martin Fischer, and Silvio Savarese.
\newblock {3D} semantic parsing of large-scale indoor spaces.
\newblock In \emph{CVPR}, 2016.

\bibitem[Ba et~al.(2016)Ba, Kiros, and Hinton]{ba2016layer}
Jimmy~Lei Ba, Jamie~Ryan Kiros, and Geoffrey~E Hinton.
\newblock Layer normalization.
\newblock \emph{arXiv preprint arXiv:1607.06450}, 2016.

\bibitem[Barr(1981)]{barr1981superquadrics}
Alan~H Barr.
\newblock Superquadrics and angle-preserving transformations.
\newblock \emph{IEEE Computer graphics and Applications}, 1981.

\bibitem[Benk{\H{o}} et~al.(2001)Benk{\H{o}}, Martin, and V{\'a}rady]{benkHo2001algorithms}
P{\'a}l Benk{\H{o}}, Ralph~R Martin, and Tam{\'a}s V{\'a}rady.
\newblock Algorithms for reverse engineering boundary representation models.
\newblock \emph{Computer-Aided Design}, 2001.

\bibitem[Binford(1975)]{binford1975visual}
Thomas Binford.
\newblock Visual perception by computer.
\newblock In \emph{Proc. IEEE Conf. on Systems and Control}, 1975.

\bibitem[Bogo et~al.(2017)Bogo, Romero, Pons-Moll, and Black]{bogo2017dynamic}
Federica Bogo, Javier Romero, Gerard Pons-Moll, and Michael~J Black.
\newblock {Dynamic FAUST: R}egistering human bodies in motion.
\newblock In \emph{CVPR}, 2017.

\bibitem[Chang et~al.(2015)Chang, Funkhouser, Guibas, Hanrahan, Huang, Li, Savarese, Savva, Song, Su, et~al.]{shapenet}
Angel~X Chang, Thomas Funkhouser, Leonidas Guibas, Pat Hanrahan, Qixing Huang, Zimo Li, Silvio Savarese, Manolis Savva, Shuran Song, Hao Su, et~al.
\newblock Shape{N}et: {A}n information-rich {3D} model repository.
\newblock \emph{arXiv preprint arXiv:1512.03012}, 2015.

\bibitem[Chen and Liu(1999)]{chen1999quadric}
YH Chen and CY Liu.
\newblock Quadric surface extraction using genetic algorithms.
\newblock \emph{Computer-Aided Design}, 1999.

\bibitem[Contributors(2022)]{spconv2022}
Spconv Contributors.
\newblock {SPConv: S}patially sparse convolution library.
\newblock \url{https://github.com/traveller59/spconv}, 2022.

\bibitem[Cui et~al.(2019)Cui, Li, Yang, Xiao, Chen, and Dong]{cui2019automatic}
Yang Cui, Qingquan Li, Bisheng Yang, Wen Xiao, Chi Chen, and Zhen Dong.
\newblock Automatic 3-d reconstruction of indoor environment with mobile laser scanning point clouds.
\newblock \emph{IEEE Journal of Selected Topics in Applied Earth Observations and Remote Sensing}, 2019.

\bibitem[Deprelle et~al.(2019)Deprelle, Groueix, Fisher, Kim, Russell, and Aubry]{deprelle2019learning}
Theo Deprelle, Thibault Groueix, Matthew Fisher, Vladimir Kim, Bryan Russell, and Mathieu Aubry.
\newblock Learning elementary structures for 3{D} shape generation and matching.
\newblock \emph{NeurIPS}, 2019.

\bibitem[Ferraz et~al.(2016)Ferraz, Saatchi, Mallet, Jacquemoud, Gon{\c{c}}alves, Silva, Soares, Tom{\'e}, and Pereira]{ferraz2016airborne}
Ant{\'o}nio Ferraz, Sassan Saatchi, Cl{\'e}ment Mallet, St{\'e}phane Jacquemoud, Gil Gon{\c{c}}alves, Carlos~Alberto Silva, Paula Soares, Margarida Tom{\'e}, and Luisa Pereira.
\newblock Airborne {LiDAR} estimation of aboveground forest biomass in the absence of field inventory.
\newblock \emph{Remote Sensing}, 2016.

\bibitem[Fischler and Bolles(1981)]{fischler1981random}
Martin~A Fischler and Robert~C Bolles.
\newblock Random sample consensus: {A} paradigm for model fitting with applications to image analysis and automated cartography.
\newblock \emph{Communications of the ACM}, 1981.

\bibitem[Gal et~al.(2009)Gal, Sorkine, Mitra, and Cohen-Or]{gal2009iwires}
Ran Gal, Olga Sorkine, Niloy~J Mitra, and Daniel Cohen-Or.
\newblock {iWIRES: A}n analyze-and-edit approach to shape manipulation.
\newblock In \emph{ACM SIGGRAPH}. 2009.

\bibitem[Gao et~al.(2021)Gao, Nan, Boom, and Ledoux]{gao2021sum}
Weixiao Gao, Liangliang Nan, Bas Boom, and Hugo Ledoux.
\newblock {SUM: A} benchmark dataset of semantic urban meshes.
\newblock \emph{ISPRS Journal of Photogrammetry and Remote Sensing}, 2021.

\bibitem[Ge et~al.(2019)Ge, Wu, Li, and Hu]{ge2019multi}
Xuming Ge, Bo Wu, Yuan Li, and Han Hu.
\newblock A multi-primitive-based hierarchical optimal approach for semantic labeling of {ALS} point clouds.
\newblock \emph{Remote Sensing}, 2019.

\bibitem[Gotardo et~al.(2003)Gotardo, Bellon, and Silva]{gotardo2003range}
Paulo~FU Gotardo, Olga Regina~Pereira Bellon, and Luciano Silva.
\newblock Range image segmentation by surface extraction using an improved robust estimator.
\newblock In \emph{CVPR}, 2003.

\bibitem[Groueix et~al.(2018)Groueix, Fisher, Kim, Russell, and Aubry]{groueix2018papier}
Thibault Groueix, Matthew Fisher, Vladimir~G Kim, Bryan~C Russell, and Mathieu Aubry.
\newblock A papier-m{\^a}ch{\'e} approach to learning {3D} surface generation.
\newblock In \emph{CVPR}, 2018.

\bibitem[Guinard et~al.(2019)Guinard, Landrieu, Caraffa, and Vallet]{guinard2019piecewise}
St{\'e}phane Guinard, Lo{\i}c Landrieu, Laurent Caraffa, and Bruno Vallet.
\newblock Piecewise-planar approximation of large 3d data as graph-structured optimization.
\newblock \emph{ISPRS Annals of Photogrammetry, Remote Sensing \& Spatial Information Sciences}, 2019.

\bibitem[Guiotte et~al.(2020)Guiotte, Pham, Dambreville, Corpetti, and Lef{\`e}vre]{guiotte2020semantic}
Florent Guiotte, Minh-Tan Pham, Romain Dambreville, Thomas Corpetti, and S{\'e}bastien Lef{\`e}vre.
\newblock Semantic segmentation of {LiDAR} points clouds: {R}asterization beyond digital elevation models.
\newblock \emph{IEEE Geoscience and Remote Sensing Letters}, 2020.

\bibitem[Guo et~al.(2016)Guo, Li, Huang, and Wang]{guo2016improved}
Bo Guo, Qingquan Li, Xianfeng Huang, and Chisheng Wang.
\newblock An improved method for power-line reconstruction from point cloud data.
\newblock \emph{Remote sensing}, 2016.

\bibitem[Gupta et~al.(2010)Gupta, Efros, and Hebert]{gupta2010blocks}
Abhinav Gupta, Alexei Efros, and Martial Hebert.
\newblock Blocks world revisited: {I}mage understanding using qualitative geometry and mechanics.
\newblock In \emph{ECCV}, 2010.

\bibitem[Haala and Brenner(1999)]{haala1999extraction}
Norbert Haala and Claus Brenner.
\newblock Extraction of buildings and trees in urban environments.
\newblock \emph{ISPRS journal of photogrammetry and remote sensing}, 1999.

\bibitem[Huang et~al.(2022)Huang, Stoter, Peters, and Nan]{rs14092254}
Jin Huang, Jantien Stoter, Ravi Peters, and Liangliang Nan.
\newblock {City3D: }large-scale building reconstruction from airborne {LiDAR} point clouds.
\newblock \emph{Remote Sensing}, 2022.

\bibitem[Ji et~al.(2019)Ji, Henriques, and Vedaldi]{ji2019invariant}
Xu Ji, Joao~F Henriques, and Andrea Vedaldi.
\newblock Invariant information clustering for unsupervised image classification and segmentation.
\newblock \emph{CVPR}, 2019.

\bibitem[Kaiser et~al.(2019)Kaiser, Ybanez~Zepeda, and Boubekeur]{kaiser2019survey}
Adrien Kaiser, Jose~Alonso Ybanez~Zepeda, and Tamy Boubekeur.
\newblock A survey of simple geometric primitives detection methods for captured {3D} data.
\newblock In \emph{Computer Graphics Forum}. Wiley Online Library, 2019.

\bibitem[Kalinicheva et~al.(2022)Kalinicheva, Landrieu, Mallet, and Chehata]{Kalinicheva_2022_CVPR}
Ekaterina Kalinicheva, Loic Landrieu, Cl\'ement Mallet, and Nesrine Chehata.
\newblock Multi-layer modeling of dense vegetation from aerial lidar scans.
\newblock In \emph{CVPR Workshops}, 2022.

\bibitem[Kingma and Ba(2014)]{kingma2014adam}
Diederik~P Kingma and Jimmy Ba.
\newblock {ADAM: A} method for stochastic optimization.
\newblock \emph{ICLR}, 2014.

\bibitem[Kluger et~al.(2021)Kluger, Ackermann, Brachmann, Yang, and Rosenhahn]{kluger2021cuboids}
Florian Kluger, Hanno Ackermann, Eric Brachmann, Michael~Ying Yang, and Bodo Rosenhahn.
\newblock Cuboids revisited: {L}earning robust {3D} shape fitting to single {RGB} images.
\newblock In \emph{CVPR}, 2021.

\bibitem[Knapitsch et~al.(2017)Knapitsch, Park, Zhou, and Koltun]{tankandtemples}
Arno Knapitsch, Jaesik Park, Qian-Yi Zhou, and Vladlen Koltun.
\newblock Tanks and temples: Benchmarking large-scale scene reconstruction.
\newblock \emph{ACM Transactions on Graphics}, 2017.

\bibitem[Kosiorek et~al.(2019)Kosiorek, Sabour, Teh, and Hinton]{kosiorek2019stacked}
Adam Kosiorek, Sara Sabour, Yee~Whye Teh, and Geoffrey~E Hinton.
\newblock Stacked capsule autoencoders.
\newblock \emph{NeurIPS}, 2019.

\bibitem[Lafarge et~al.(2006)Lafarge, Descombes, Zerubia, and Pierrot-Deseilligny]{lafarge2006automatic}
Florent Lafarge, Xavier Descombes, Josiane Zerubia, and Marc Pierrot-Deseilligny.
\newblock Automatic 3{D} builfding reconstruction from {DEMs: A}n application to {PLEIADES} simulations.
\newblock In \emph{ISPRS}, 2006.

\bibitem[Leonardis et~al.(1995)Leonardis, Gupta, and Bajcsy]{leonardis1995segmentation}
Ale{\v{s}} Leonardis, Alok Gupta, and Ruzena Bajcsy.
\newblock Segmentation of range images as the search for geometric parametric models.
\newblock \emph{International Journal of Computer Vision}, 1995.

\bibitem[Leonardis et~al.(1997)Leonardis, Jaklic, and Solina]{leonardis1997superquadrics}
Ales Leonardis, Ales Jaklic, and Franc Solina.
\newblock Superquadrics for segmenting and modeling range data.
\newblock \emph{IEEE Transactions on Pattern Analysis and Machine Intelligence}, 1997.

\bibitem[Li et~al.(2017)Li, Xu, Chaudhuri, Yumer, Zhang, and Guibas]{li2017grass}
Jun Li, Kai Xu, Siddhartha Chaudhuri, Ersin Yumer, Hao Zhang, and Leonidas Guibas.
\newblock Grass: {G}enerative recursive autoencoders for shape structures.
\newblock \emph{ACM Transactions on Graphics}, 2017.

\bibitem[Li et~al.(2019)Li, Sung, Dubrovina, Yi, and Guibas]{li2019supervised}
Lingxiao Li, Minhyuk Sung, Anastasia Dubrovina, Li Yi, and Leonidas~J Guibas.
\newblock Supervised fitting of geometric primitives to {3D} point clouds.
\newblock In \emph{CVPR}, 2019.

\bibitem[Li et~al.(2016)Li, Wonka, and Nan]{li2016manhattan}
Minglei Li, Peter Wonka, and Liangliang Nan.
\newblock Manhattan-world urban reconstruction from point clouds.
\newblock In \emph{ECCV}, 2016.

\bibitem[Li et~al.(2011)Li, Wu, Chrysathou, Sharf, Cohen-Or, and Mitra]{li2011globfit}
Yangyan Li, Xiaokun Wu, Yiorgos Chrysathou, Andrei Sharf, Daniel Cohen-Or, and Niloy~J Mitra.
\newblock Globfit: {C}onsistently fitting primitives by discovering global relations.
\newblock In \emph{ACM SIGGRAPH}. 2011.

\bibitem[Liao et~al.(2022)Liao, Xie, and Geiger]{kitti360}
Yiyi Liao, Jun Xie, and Andreas Geiger.
\newblock {KITTI-360}: {A} novel dataset and benchmarks for urban scene understanding in {2D} and {3D}.
\newblock \emph{IEEE Transactions on Pattern Analysis and Machine Intelligence}, 2022.

\bibitem[Liu et~al.(2022)Liu, Wu, Ruan, and Chirikjian]{liu2022robust}
Weixiao Liu, Yuwei Wu, Sipu Ruan, and Gregory~S Chirikjian.
\newblock Robust and accurate superquadric recovery: {A} probabilistic approach.
\newblock In \emph{CVPR}, 2022.

\bibitem[Locatello et~al.(2020)Locatello, Weissenborn, Unterthiner, Mahendran, Heigold, Uszkoreit, Dosovitskiy, and Kipf]{locatello2020object}
Francesco Locatello, Dirk Weissenborn, Thomas Unterthiner, Aravindh Mahendran, Georg Heigold, Jakob Uszkoreit, Alexey Dosovitskiy, and Thomas Kipf.
\newblock Object-centric learning with slot attention.
\newblock \emph{NeurIPS}, 2020.

\bibitem[Lodha et~al.(2006)Lodha, Kreps, Helmbold, and Fitzpatrick]{lodha2006aerial}
Suresh~K Lodha, Edward~J Kreps, David~P Helmbold, and Darren Fitzpatrick.
\newblock Aerial {LiDAR} data classification using support vector machines.
\newblock In \emph{Third International Symposium on {3D} Data Processing, Visualization, and Transmission (3DPVT'06)}. IEEE, 2006.

\bibitem[Loiseau et~al.(2021)Loiseau, Monnier, Aubry, and Landrieu]{loiseau2021representing}
Romain Loiseau, Tom Monnier, Mathieu Aubry, and Lo{\"\i}c Landrieu.
\newblock Representing shape collections with alignment-aware linear models.
\newblock In \emph{3DV}, 2021.

\bibitem[MacQueen(1967)]{kmeans}
J MacQueen.
\newblock Classification and analysis of multivariate observations.
\newblock In \emph{5th Berkeley Symp. Math. Statist. Probability}, 1967.

\bibitem[mapping agency)(2021)]{lidarhd}
IGN~(French mapping agency).
\newblock {LiDAR-HD: A {3D} mapping of France’s soil and subsoil}, 2021.

\bibitem[McCormac et~al.(2017)McCormac, Handa, Leutenegger, and J.Davison]{McCormac2017Scene}
John McCormac, Ankur Handa, Stefan Leutenegger, and Andrew J.Davison.
\newblock {SceneNet RGB-D: C}an 5{M} synthetic images beat generic imagenet pre-training on indoor segmentation?
\newblock \emph{ICCV}, 2017.

\bibitem[Mei et~al.(2017)Mei, Zhang, Wu, Wang, and Zhang]{mei20173d}
Jie Mei, Liqiang Zhang, Shihao Wu, Zhen Wang, and Liang Zhang.
\newblock 3d tree modeling from incomplete point clouds via optimization and l 1-mst.
\newblock \emph{International Journal of Geographical Information Science}, 2017.

\bibitem[Monnier et~al.(2020)Monnier, Groueix, and Aubry]{monnier2020dticlustering}
Tom Monnier, Thibault Groueix, and Mathieu Aubry.
\newblock {Deep Transformation-Invariant Clustering}.
\newblock \emph{NeurIPS}, 2020.

\bibitem[Monnier et~al.(2021)Monnier, Vincent, Ponce, and Aubry]{monnier2021unsupervised}
Tom Monnier, Elliot Vincent, Jean Ponce, and Mathieu Aubry.
\newblock Unsupervised layered image decomposition into object prototypes.
\newblock In \emph{ICCV}, 2021.

\bibitem[Morsdorf et~al.(2004)Morsdorf, Meier, K{\"o}tz, Itten, Dobbertin, and Allg{\"o}wer]{morsdorf2004lidar}
Felix Morsdorf, Erich Meier, Benjamin K{\"o}tz, Klaus~I Itten, Matthias Dobbertin, and Britta Allg{\"o}wer.
\newblock {LIDAR}-based geometric reconstruction of boreal type forest stands at single tree level for forest and wildland fire management.
\newblock \emph{Remote sensing of environment}, 2004.

\bibitem[Nan and Wonka(2017)]{nan2017polyfit}
Liangliang Nan and Peter Wonka.
\newblock Polyfit: {P}olygonal surface reconstruction from point clouds.
\newblock In \emph{ICCV}, 2017.

\bibitem[Niemeyer et~al.(2014)Niemeyer, Rottensteiner, and Soergel]{niemeyer2014contextual}
Joachim Niemeyer, Franz Rottensteiner, and Uwe Soergel.
\newblock Contextual classification of {LiDAR} data and building object detection in urban areas.
\newblock \emph{ISPRS journal of photogrammetry and remote sensing}, 2014.

\bibitem[Paschalidou et~al.(2019)Paschalidou, Ulusoy, and Geiger]{paschalidou2019superquadrics}
Despoina Paschalidou, Ali~Osman Ulusoy, and Andreas Geiger.
\newblock Superquadrics revisited: {L}earning {3D} shape parsing beyond cuboids.
\newblock In \emph{CVPR}, 2019.

\bibitem[Paschalidou et~al.(2020)Paschalidou, Gool, and Geiger]{paschalidou2020learning}
Despoina Paschalidou, Luc~Van Gool, and Andreas Geiger.
\newblock Learning unsupervised hierarchical part decomposition of {3D} objects from a single {RGB} image.
\newblock In \emph{CVPR}, 2020.

\bibitem[Paschalidou et~al.(2021)Paschalidou, Katharopoulos, Geiger, and Fidler]{paschalidou2021neural}
Despoina Paschalidou, Angelos Katharopoulos, Andreas Geiger, and Sanja Fidler.
\newblock Neural parts: {L}earning expressive {3D} shape abstractions with invertible neural networks.
\newblock In \emph{CVPR}, 2021.

\bibitem[Ramamonjisoa et~al.(2022)Ramamonjisoa, Stekovic, and Lepetit]{ramamonjisoa2022monteboxfinder}
Micha{\"e}l Ramamonjisoa, Sinisa Stekovic, and Vincent Lepetit.
\newblock Monteboxfinder: {D}etecting and filtering primitives to fit a noisy point cloud.
\newblock In \emph{ECCV}. Springer, 2022.

\bibitem[Ravi et~al.(2020)Ravi, Reizenstein, Novotny, Gordon, Lo, Johnson, and Gkioxari]{ravi2020pytorch3d}
Nikhila Ravi, Jeremy Reizenstein, David Novotny, Taylor Gordon, Wan-Yen Lo, Justin Johnson, and Georgia Gkioxari.
\newblock Accelerating 3{D} deep learning with {PyTorch3D}.
\newblock \emph{arXiv:2007.08501}, 2020.

\bibitem[Roberts(1963)]{roberts1963machine}
Lawrence~G. Roberts.
\newblock \emph{Machine perception of three-dimensional solids}.
\newblock PhD thesis, Massachusetts Institute of Technology, 1963.

\bibitem[Savva et~al.(2015{\natexlab{a}})Savva, Chang, and Hanrahan]{savva2015semgeo}
Manolis Savva, Angel~X. Chang, and Pat Hanrahan.
\newblock {Semantically-Enriched {3D} Models for Common-sense Knowledge}.
\newblock \emph{CVPR Workshop}, 2015{\natexlab{a}}.

\bibitem[Savva et~al.(2015{\natexlab{b}})Savva, Chang, and Hanrahan]{shapenetsem}
Manolis Savva, Angel~X. Chang, and Pat Hanrahan.
\newblock {Semantically-Enriched {3D} Models for Common-sense Knowledge}.
\newblock \emph{CVPR 2015 Workshop on Functionality, Physics, Intentionality and Causality}, 2015{\natexlab{b}}.

\bibitem[Schnabel et~al.(2007)Schnabel, Wahl, and Klein]{schnabel2007efficient}
Ruwen Schnabel, Roland Wahl, and Reinhard Klein.
\newblock Efficient ransac for point-cloud shape detection.
\newblock In \emph{Computer Graphics Forum}, 2007.

\bibitem[Schnabel et~al.(2009)Schnabel, Degener, and Klein]{schnabel2009completion}
Ruwen Schnabel, Patrick Degener, and Reinhard Klein.
\newblock Completion and reconstruction with primitive shapes.
\newblock In \emph{Computer Graphics Forum}. Wiley Online Library, 2009.

\bibitem[Sharma et~al.(2018)Sharma, Goyal, Liu, Kalogerakis, and Maji]{sharma2018csgnet}
Gopal Sharma, Rishabh Goyal, Difan Liu, Evangelos Kalogerakis, and Subhransu Maji.
\newblock {CSGNet: N}eural shape parser for constructive solid geometry.
\newblock In \emph{CVPR}, 2018.

\bibitem[Singer and Asari(2021)]{singer2021dales}
Nina~M Singer and Vijayan~K Asari.
\newblock {DALES} objects: {A} large scale benchmark dataset for instance segmentation in aerial {LiDAR}.
\newblock \emph{IEEE Access}, 2021.

\bibitem[Smirnov et~al.(2021)Smirnov, Gharbi, Fisher, Guizilini, Efros, and Solomon]{smirnov_marionette_2021}
Dmitriy Smirnov, Michael Gharbi, Matthew Fisher, Vitor Guizilini, Alexei Efros, and Justin~M Solomon.
\newblock Marionette: {S}elf-supervised sprite learning.
\newblock \emph{NeurIPS}, 2021.

\bibitem[Sun et~al.(2022)Sun, Huang, Zhang, Chen, An, Wang, and Yun]{sun2022individual}
Chenxin Sun, Chengwei Huang, Huaiqing Zhang, Bangqian Chen, Feng An, Liwen Wang, and Ting Yun.
\newblock Individual tree crown segmentation and crown width extraction from a heightmap derived from aerial laser scanning data using a deep learning framework.
\newblock \emph{Frontiers in plant science}, 2022.

\bibitem[Sung et~al.(2015)Sung, Kim, Angst, and Guibas]{sung2015data}
Minhyuk Sung, Vladimir~G Kim, Roland Angst, and Leonidas Guibas.
\newblock Data-driven structural priors for shape completion.
\newblock \emph{ACM Transactions on Graphics}, 2015.

\bibitem[Tulsiani et~al.(2017)Tulsiani, Su, Guibas, Efros, and Malik]{tulsiani2017learning}
Shubham Tulsiani, Hao Su, Leonidas~J Guibas, Alexei~A Efros, and Jitendra Malik.
\newblock Learning shape abstractions by assembling volumetric primitives.
\newblock In \emph{CVPR}, 2017.

\bibitem[Varney et~al.(2020)Varney, Asari, and Graehling]{varney2020dales}
Nina Varney, Vijayan~K Asari, and Quinn Graehling.
\newblock {DALES: A} large-scale aerial {LiDAR} data set for semantic segmentation.
\newblock In \emph{CVPR Workshop}, 2020.

\bibitem[Vauhkonen et~al.(2012)Vauhkonen, Ene, Gupta, Heinzel, Holmgren, Pitk{\"a}nen, Solberg, Wang, Weinacker, Hauglin, et~al.]{vauhkonen2012comparative}
Jari Vauhkonen, Liviu Ene, Sandeep Gupta, Johannes Heinzel, Johan Holmgren, Juho Pitk{\"a}nen, Svein Solberg, Yunsheng Wang, Holger Weinacker, K~Marius Hauglin, et~al.
\newblock Comparative testing of single-tree detection algorithms under different types of forest.
\newblock \emph{Forestry}, 2012.

\bibitem[Weinstein et~al.(2019)Weinstein, Marconi, Bohlman, Zare, and White]{weinstein2019individual}
Ben~G Weinstein, Sergio Marconi, Stephanie Bohlman, Alina Zare, and Ethan White.
\newblock Individual tree-crown detection in {RGB} imagery using semi-supervised deep learning neural networks.
\newblock \emph{Remote Sensing}, 2019.

\bibitem[Wu et~al.(2022)Wu, Liu, Ruan, and Chirikjian]{wu2022primitive}
Yuwei Wu, Weixiao Liu, Sipu Ruan, and Gregory~S Chirikjian.
\newblock Primitive-based shape abstraction via nonparametric bayesian inference.
\newblock In \emph{ECCV}, 2022.

\bibitem[Wu et~al.(2015)Wu, Song, Khosla, Yu, Zhang, Tang, and Xiao]{modelnet}
Zhirong Wu, Shuran Song, Aditya Khosla, Fisher Yu, Linguang Zhang, Xiaoou Tang, and Jianxiong Xiao.
\newblock {3D ShapeNets: A} deep representation for volumetric shapes.
\newblock In \emph{CVPR}, 2015.

\bibitem[Xia et~al.(2020)Xia, Chen, Wang, Li, and Zhang]{xia2020geometric}
Shaobo Xia, Dong Chen, Ruisheng Wang, Jonathan Li, and Xinchang Zhang.
\newblock Geometric primitives in {LiDAR} point clouds: {A} review.
\newblock \emph{IEEE Journal of Selected Topics in Applied Earth Observations and Remote Sensing}, 2020.

\bibitem[Ye et~al.(2020)Ye, Xu, Huang, Tong, Li, Liu, Luan, Hoegner, and Stilla]{ye2020lasdu}
Zhen Ye, Yusheng Xu, Rong Huang, Xiaohua Tong, Xin Li, Xiangfeng Liu, Kuifeng Luan, Ludwig Hoegner, and Uwe Stilla.
\newblock {LASDU: A} large-scale aerial {LiDAR} dataset for semantic labeling in dense urban areas.
\newblock \emph{ISPRS International Journal of Geo-Information}, 2020.

\bibitem[Yu et~al.(2022)Yu, Guibas, and Wu]{yuUnsupervisedDiscoveryObject2021}
Hong-Xing Yu, Leonidas~J. Guibas, and Jiajun Wu.
\newblock Unsupervised {{Discovery}} of {{Object Radiance Fields}}.
\newblock \emph{ICLR}, 2022.

\bibitem[Zhang et~al.(2021{\natexlab{a}})Zhang, Ti{\v{n}}o, Leonardis, and Tang]{zhang2021survey}
Yu Zhang, Peter Ti{\v{n}}o, Ale{\v{s}} Leonardis, and Ke Tang.
\newblock A survey on neural network interpretability.
\newblock \emph{IEEE Transactions on Emerging Topics in Computational Intelligence}, 2021{\natexlab{a}}.

\bibitem[Zhang et~al.(2021{\natexlab{b}})Zhang, Girdhar, Joulin, and Misra]{Zhang_2021_ICCV}
Zaiwei Zhang, Rohit Girdhar, Armand Joulin, and Ishan Misra.
\newblock Self-supervised pretraining of 3d features on any point-cloud.
\newblock In \emph{ICCV}, 2021{\natexlab{b}}.

\bibitem[Zhao et~al.(2019)Zhao, Birdal, Deng, and Tombari]{zhao20193d}
Yongheng Zhao, Tolga Birdal, Haowen Deng, and Federico Tombari.
\newblock {3D} point capsule networks.
\newblock In \emph{CVPR}, 2019.

\bibitem[Zolanvari et~al.(2019)Zolanvari, Ruano, Rana, Cummins, da~Silva, Rahbar, and Smolic]{zolanvari2019dublincity}
SM Zolanvari, Susana Ruano, Aakanksha Rana, Alan Cummins, Rogerio~Eduardo da Silva, Morteza Rahbar, and Aljosa Smolic.
\newblock {DublinCity:} annotated {LiDAR} point cloud and its applications.
\newblock \emph{BMVC}, 2019.

\bibitem[Zou et~al.(2017)Zou, Yumer, Yang, Ceylan, and Hoiem]{zou20173d}
Chuhang Zou, Ersin Yumer, Jimei Yang, Duygu Ceylan, and Derek Hoiem.
\newblock {3D-PRNN}: {G}enerating shape primitives with recurrent neural networks.
\newblock In \emph{ICCV}, 2017.

\end{thebibliography}
}

\clearpage
\section{Supplementary Material}

In this supplementary material, we provide details on the implementation of our \METHODNAME~(\Secref{sec:suppmat:implem}), details about our proposed \DATASETNAME~(\Secref{sec:suppmat:dataset}), and some additional quantitative~(\Secref{sec:suppmat:quanti}) and qualitative~(\Secref{sec:suppmat:quali}) results. Our code and dataset are available at \url{https://romainloiseau.fr/learnable-earth-parser/}.
\section{Detailed Configuration}
\label{sec:suppmat:implem}

\begin{figure*}
    \includegraphics[trim={.45cm 0 .45cm 0},clip,width=\textwidth]{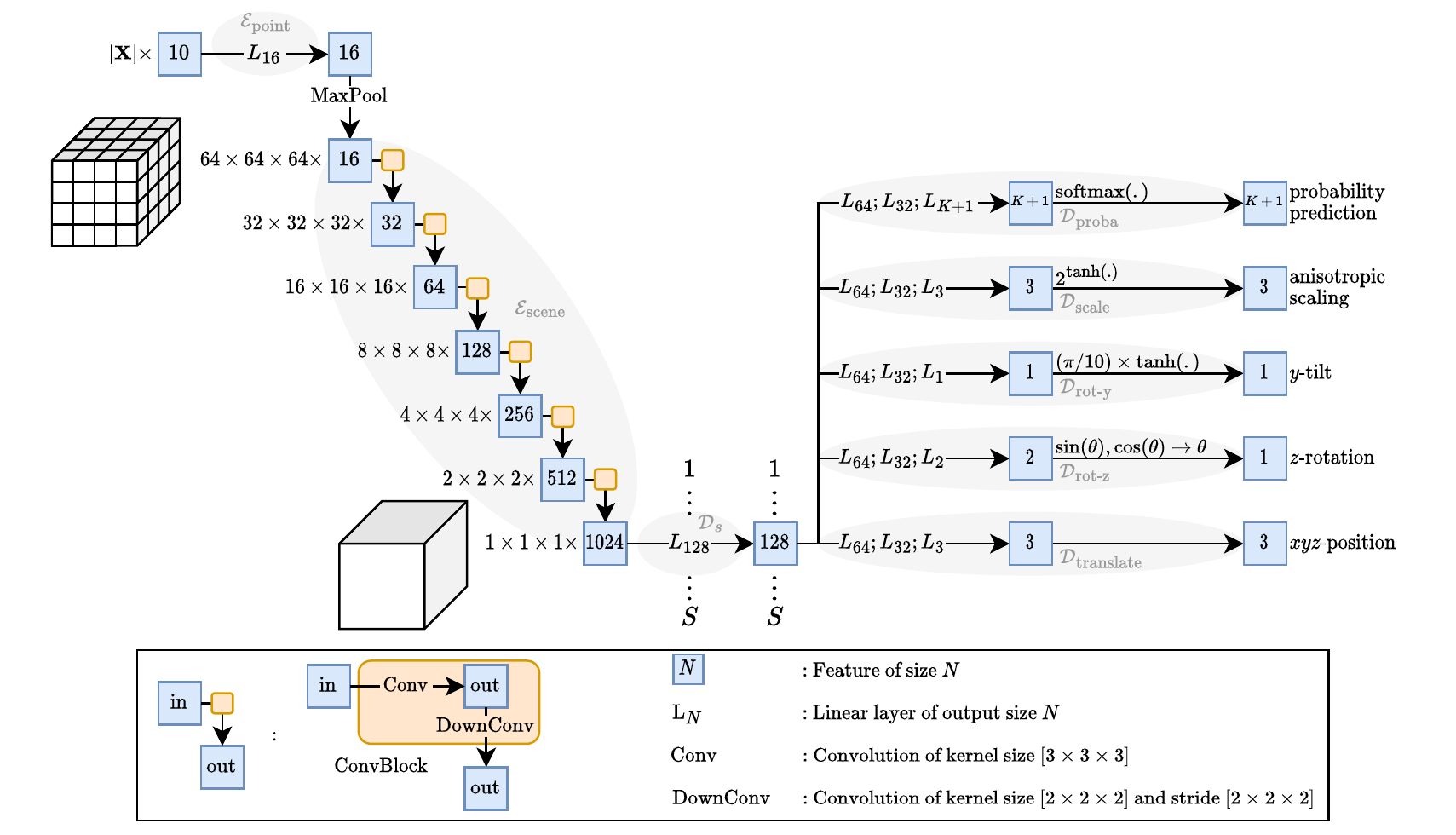}
    \caption{\textbf{\METHODNAME~Detailed Architecture}. Details of the architecture showing all layers in $\cE_\text{point}, \cE_\text{scene}, \cD_s, \cD_\text{proba}, \cD_\text{scale}, \cD_\text{rot-y}, \cD_\text{rot-z}$ and $\cD_\text{translate}$. We use LayerNorm~\cite{ba2016layer} and LeakyRelu after all hidden layers.}\label{fig:network}
\end{figure*}

We report here the exact architecture of the \METHODNAME \ network and training details.

\ROM{\paragraph{Learnable prototypes.}Following Loiseau~\etal~\cite{loiseau2021representing}, the point coordinates of our prototypes $\proto^1,\cdots,\proto^K$ are learned directly as free parameters of the model through our reconstruction loss. Each prototype contains $256$ points leading to learning $K\times256\times3$ free parameters to represent all the learned prototypes.} Eventually, our model's 3D prototypes are defined by their points’ coordinates, which are free parameters learned by optimizing the reconstruction loss $\mathcal{L}_\text{rec}$ and its regularization $\mathcal{L}_\text{reg}$. 
While the reconstruction task serves as a label-free supervisory signal, our main goal is not to achieve the best possible reconstruction but to learn simple and interpretable prototypes. A model using feature-space prototypes and arbitrary transformations may achieve a much lower reconstruction error, but its prototypes would have low semantic purity and interpretability. 

\paragraph{Network architecture.}
Our model takes a point cloud $\bX$ and computes a voxelization in a grid of size $64 \times 64  \times 64$. As shown in~\figref{fig:network}, our model is composed of (i) a point encoder $\cE_\text{point}$,
(ii) a scene encoder $\cE_\text{scene}$,
(iii) $S$ slot feature extractors $\cD_s$ and (iv) five shared slot parameters generators: $\cD_\text{proba}, \cD_\text{scale}, \cD_\text{rot-y}, \cD_\text{rot-z}, \cD_\text{translate}$. 
We provide details on these networks below.
\begin{itemize}[itemsep=0em, wide, labelwidth=!, labelindent=0pt]
    \item \textbf{Point encoder.}~Each input point of $\bX$ is associated with a 10-dimensional descriptor: (1-3) normalized position in the tile in $[-1,1]^3$, (4-6) \emph{rgb} color, (7) normalized LiDAR reflectance, and (8-10)  its offset relative to the center of its assigned voxel. The point encoder $\cE_\text{point}$ is a linear layer that maps these descriptors to a 16-dimensional point feature.
    \item \textbf{Scene encoder.}~We compute voxel features by max-pooling the features of the points associated to each voxel. The scene encoder $\cE_\text{scene}$ then maps these voxel features to a single scene feature, a vector of size $1024$, by using a sequence of $6$ 3D sparse convolutions~\cite{spconv2022} with kernel size $[3,3,3]$ and $6$ strided convolutions with kernel size $[2,2,2]$ and stride $[2,2,2]$.
    \item \textbf{Slot feature extractor.}~
    Each slot $s$ takes as input the scene feature produced by $\cE_\text{scene}$ and maps it to a slot feature of size $128$ with a dedicated linear layer $\cD_s$.
    \item \textbf{Slot parameters generators.}~
    Five 3-layers Multi Layer Perceptrons (MLPs) are shared by all slots to map their slot features to the associated parameters of the reconstruction model.
    \begin{itemize}
        \item[-] $\cD_\text{proba}$ outputs the slot activation and prototype choice probability $\alpha_s$ et $\beta^k_s$.
         \item[-] $\cD_\text{scale}$ outputs three scales in $[-1/2,2]$, corresponding to scaling the prototypes in each canonical directions.
         \item[-] $\cD_\text{rot-y}$ outputs a rotation in $[-\pi/10,\pi/10]$ to be applied around the $y$ axis.
        \item[-] $\cD_\text{rot-z}$  outputs a 2D point on the unit circle which is then mapped to a rotation in $[-\pi,\pi]$ to be applied around the $z$ axis.
         \item[-] $\cD_\text{translate}$ outputs a 3D translation vector in $\mathbb{R}^3$.
    \end{itemize}
     These parameters are used to determine the activation of the slot, choose a prototype, then apply a sequence of transformations in the following order:  scaling, $y$-rotation, $z$-rotation, and translation.
\end{itemize}

\paragraph{Reconstruction loss.}Due to the arbitrary square shape of our samples $\bX$, some objects can appear only partly in a patch. We don't want the network to learn prototypes specifically to fit such object parts, as it is an artifact of our sampling procedure. Indeed, square patches are sampled randomly during training, and along a non-overlapping grid for inference. Instead, we propose to ignore the points of the reconstruction $\bY_s^k$ that falls beyond the normalized $[-1,1]$ extent of the patches. This allows the network to predict full objects without being penalized in terms of accuracy. To do so, we modify Equation \cy{8} from the main paper as follows:
\begin{align}
    \cL_\text{acc}(\cM,\bX)=\frac{1}{S}\sum_{s=1}^S\sum_{k=1}^K{\beta_{s}^{k}}{ d\left(\widetilde{\bY}_s^k,\bX\right)}~,
\end{align}
where $\widetilde{\bY}_s^k$ is the subset of points of $\bY_s^k$ that falls within the horizontal extent of their patch $[-1,1]^2 \times \eR$.
To prevent the slots from predicting shapes outside of the patch extent, we regularize our model by the square Euclidean distance between the output of $\mathcal{D}_\text{translate}$ and the set $[-1,1]^2\times\eR$ for each slot.

\paragraph{Training.}We use the efficient CUDA implementation of the Chamfer distance by PyTorch3D~\cite{ravi2020pytorch3d} which significantly speeds up training. We use the ADAM optimizer~\cite{kingma2014adam} with a learning rate of $10^{-4}$ and default parameters, except for the prototypes' intensities, scales and points' positions which we learn without weight decay.

\paragraph{Curriculum learning.}Following the ideas of  Monnier~\etal~\cite{monnier2020dticlustering} and Loiseau~\etal~\cite{loiseau2021representing}, we use a multi-stage curriculum strategy to prevent our model from falling in bad minima. We gradually unfreeze the model parameters in the following order: (i) translation, rotation, tilt, slot activation, and choice of prototype; (ii) intensities of the prototypes, when available; (iii) scales of the prototypes; (iv) shapes of the prototypes (positions of their 3D points); (v) anisotropic scalings of the prototypes. Alignment networks are initially set to identity by setting the parameters of the decoders' last linear layers to zero. When unfreezing a new module, the learning rate of all the model's parameters is set to ${1}/{1000}$ of the global learning rate and gradually increased over $1000$ batches to the global learning rate to smooth the training and benefit from what has been learned previously by the encoder. We define an ``epoch" as $512$ batches of $64$ patches, and each stage of the curriculum is trained until convergence.

\paragraph{Scene-specific hyperparameters.}We trained our model on each seven scenes of the \DATASETNAME, with minor adaptation shown in \tabref{tab:adapt}. We only change the size of the voxel grid to adapt our reconstruction model to the size of the typical object we want to discover. For example, a windturbine is typically $100$ meters tall, while a boat is typically $5$ meters long. We also doubled the number of slots for the ``\powerplant" scene because of its geometric complexity.

\begin{table}[t]
    \centering
    \caption{\textbf{\METHODNAME \ hyperparameters.}~Choice of hyperparameters when training on the \DATASETNAME. We used similar configurations across scenes, only adapting the voxel size and number of slots.}
    \begin{tabular}{@{}lcc@{}}\toprule
        Scene           & Voxel size (cm)  & number of slots $S$    \\\midrule
        \cropfield	    & $40$	   & $64$        \\
        \forest         & $60$	   & $64$                   \\
        \greenhouse	    & $60$	   & $64$\\
        \boats	        & $20$	   & $64$            \\
        \powerplant	    & $60$	   & $128$         \\
        \nimes	        & $40$	   & $64$       \\
        \windturbine	& $320$	   & $64$             \\
        \bottomrule
    \end{tabular}
    \label{tab:adapt}
\end{table}

\paragraph{ShapeNet adaptations.}
As the objects of ShapeNet-Part~\cite{shapenetsem} are simple, we only use $S=6$ slots. To account for the  diversity of the size and shapes of parts, we replaced the anisotropic-scaling transformation by an unconstrained affine transformation.

\paragraph{Ablation Study.}The structure of our ablation study intentionally mirrors the curriculum learning. Specifically, we remove components in decreasing order of their impact on the reconstruction quality. Removing the translation while retaining all other transformations would lead to poor {learning dynamics~\cite{monnier2020dticlustering}}. Our approach ensures that each step of the ablation progressively assesses the impact of each component.
\section{\DATASETNAME~Details}\label{sec:suppmat:dataset}

\paragraph{Classes names.}As show in \tabref{tab:dataset:details}, each scene of the \DATASETNAME \ is annotated with different classes among ``ground", ``vegetation", ``building", ``boats", ``bridge", ``electric lines", and ``windturbine".

\paragraph{Localization.}We report the localization of the scenes of \DATASETNAME \ in \tabref{tab:dataset:details}. Our dataset has been acquired in various environments distributed on the French territory.

\begin{table*}[p]
    \centering
    \caption{\textbf{\DATASETNAME~classes and localisation.}~We show the class names and color codes for the seven scenes of our dataset. Unlabeled points are represented in black~\protect\cbox{unlabeled}~. The \DATASETNAME \ was acquired at different locations in France, spanning a wide variety of environments.}
    \begin{tabular}{@{}r@{~}lr@{~}lr@{~}lr@{~}lr@{~}lr@{~}lr@{~}l@{}}
    \toprule
         \includegraphics[width=.3cm]{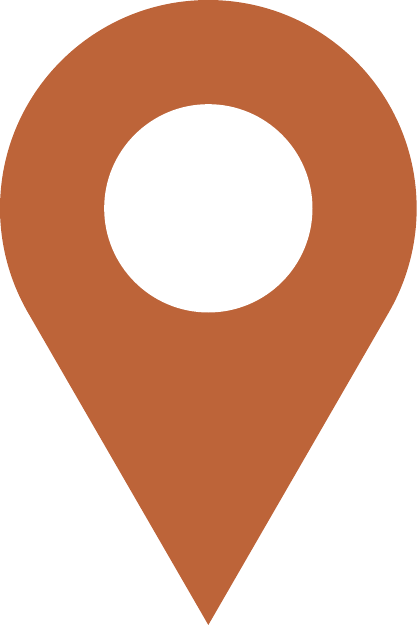} & \cropfield &
         \includegraphics[width=.3cm]{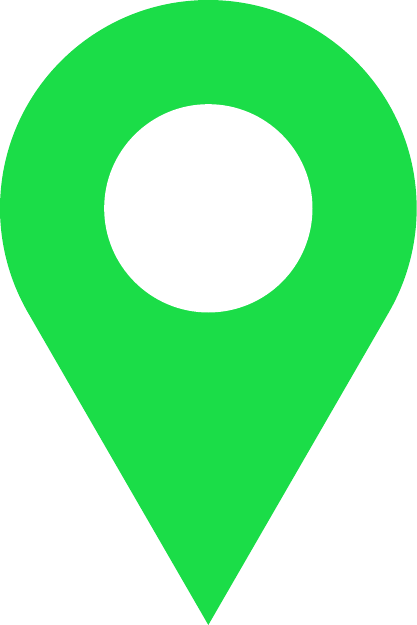} & \forest &
         \includegraphics[width=.3cm]{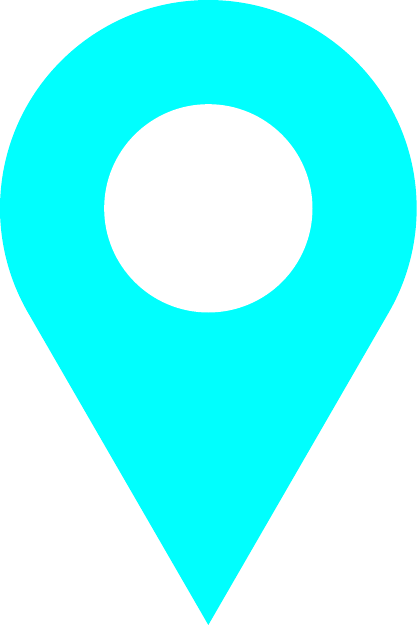} & \greenhouse &
         \includegraphics[width=.3cm]{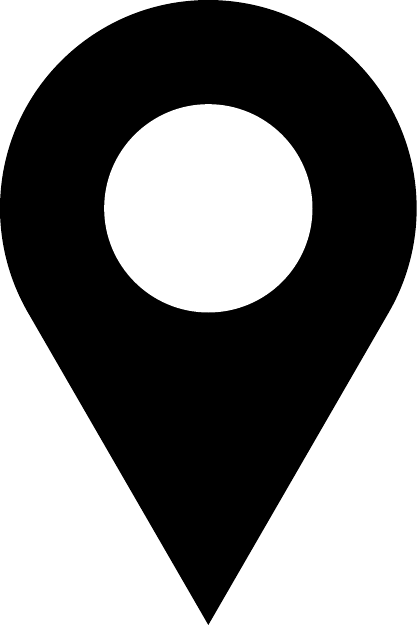} & \boats &
         \includegraphics[width=.3cm]{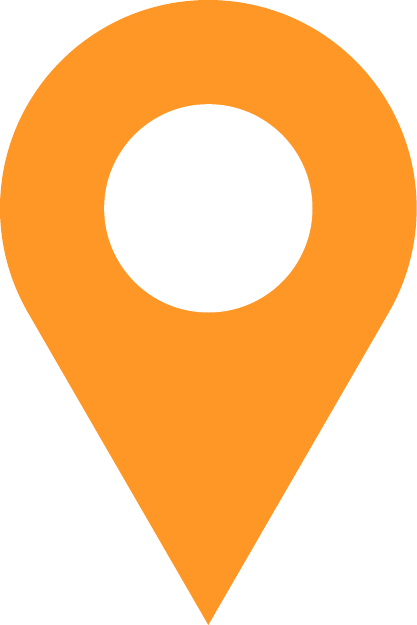} & \powerplant &
         \includegraphics[width=.3cm]{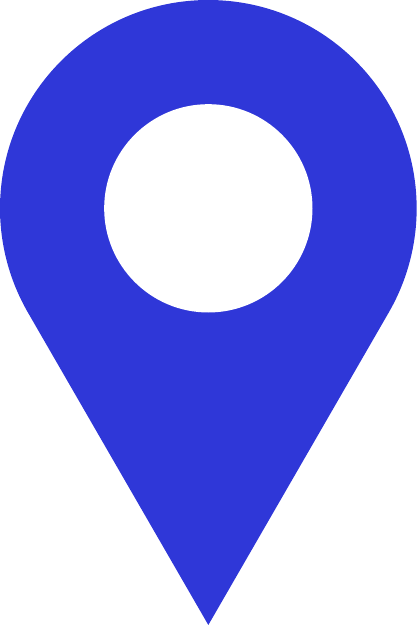} & \nimes &
         \includegraphics[width=.3cm]{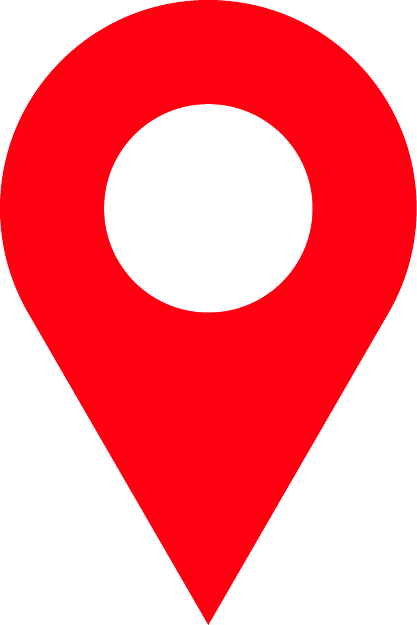} & \windturbine
         \\\midrule
         \cboxlbl{ground}{Ground} & \cboxlbl{ground}{Ground} &\cboxlbl{ground}{Ground}&\cboxlbl{boats}{Boats} & \cboxlbl{ground}{Ground}&\cboxlbl{ground}{Ground}&\cboxlbl{ground}{Ground}\\
         \cboxlbl{vegetation}{Vegetation} &\cboxlbl{vegetation}{Vegetation} &\cboxlbl{vegetation}{Vegetation}&\cboxlbl{bridge!90!white}{Bridge} &\cboxlbl{vegetation}{Vegetation} &\cboxlbl{vegetation}{Vegetation} &\cboxlbl{vegetation}{Vegetation}  \\
         && && \cboxlbl{building}{Building} &&& \cboxlbl{building}{Building} & \cboxlbl{building}{Building} & \cboxlbl{windturbine}{Windturbine}\\
         && && && && \cboxlbl{electric}{Electric lines}\\\midrule
         \multicolumn{14}{@{}c@{}}{\includegraphics[width=.8\linewidth]{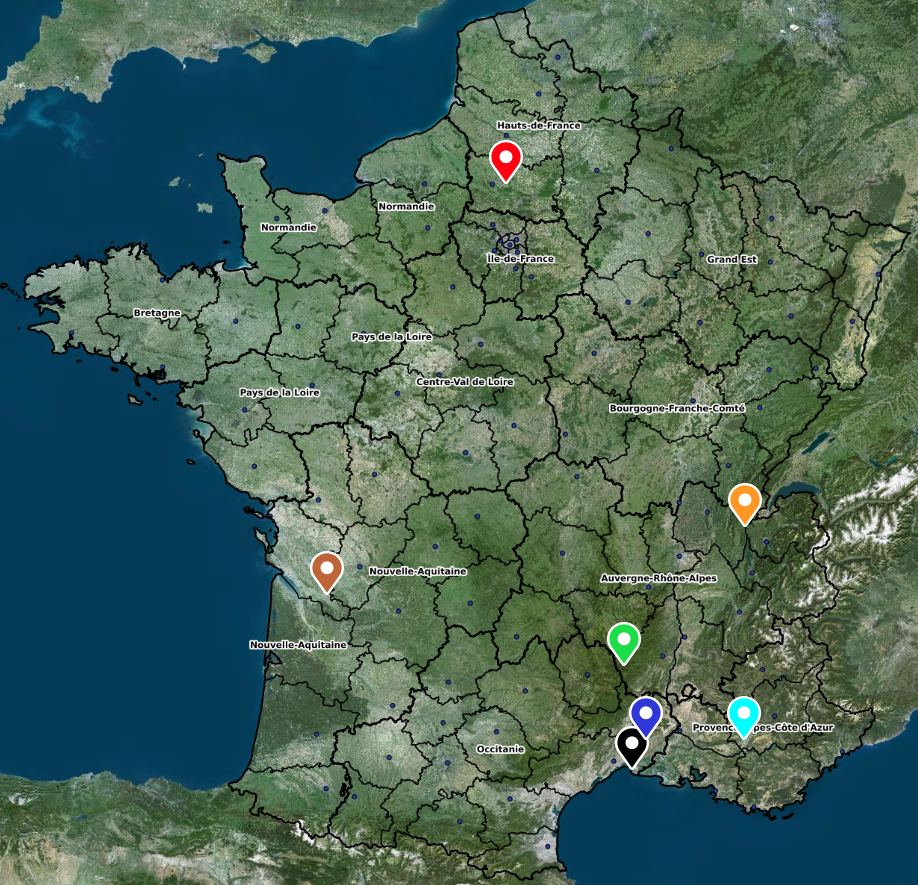}}
         \\\bottomrule
    \end{tabular}
    \label{tab:dataset:details}
\end{table*}
\section{Additional Quantitative Results}\label{sec:suppmat:quanti}

\paragraph{Results on the \DATASETNAME.}We report in \tabref{tab:suppmat:quantidetails} detailed results for the baselines and our method. We evaluated the use of elevation and LiDAR intensity for the k-means~\cite{kmeans} baseline, the use of intensity in a way similar to ours for AtlasNet~v2~\cite{deprelle2019learning}, and the effect of our prototype selection post-processing step:
\begin{itemize}[itemsep=0em, wide, labelwidth=!, labelindent=0pt]
\item \textbf{k-means features.}~The use of both intensity and elevation gives a small boost to semantic performances. However, we see that when clustering with a small number of centroids ($K=6$, as in our model), using only the elevation gives a reasonable baseline.
\item \textbf{AtlasNet~v2 intensity.}~We extend AtlasNet~v2 to handle intensity in a manner similar to our approach, which improves its segmentation results.  However, AtlasNet~v2 uses the same number of prototypes for each input regardless of its complexity and thus does not achieve high semantic segmentation scores.
\item \textbf{Prototype selection post-processing.}~On the \METHODNAME, we see that our post-processing step has a limited impact on the quality of the prediction and reconstructions, except for the scene ``\forest" for which the segmentation score goes from $87.3$ to $80.5$. This step can either increase (``\cropfield", ``\forest", ``\boats" scenes) or decrease (``\windturbine" scene) the reconstruction quality. We believe this is because of the regularization loss which encourages all prototypes to be used. Finally, this simple post-processing step allows us to significantly decrease the number of prototypes and adapt it to the scene complexity.
\end{itemize}

\ROM{\paragraph{Results on ShapeNet.}Our experiment on ShapeNet is intended as a sanity check in a controlled setting. We report in Table~\cy{4} significantly better results than AtlasNet\,v2~\cite{deprelle2019learning} for planes with arbitrary orientation ($+34.1$ mIoU). We repeated the experiment for guitars and chairs and observed improvements of $+23.1$ and $+1.5$ mIoU, respectively.}

\ROM{\paragraph{Additional ablations.}The coordinates of the prototypes' points are directly learned as parameters of the model in an unsupervised fashion with our regularized reconstruction loss. We choose point cloud prototypes for their simplicity and expressivity. We also trained our model using cuboids or superquadrics as prototypes, by learning their parameters ($xyz$-scales for cuboids, and $xyz$-scales and $\alpha_1,\alpha_2$ for superquadrics) as free parameters of the model for each prototypes. This leads to worse reconstruction results (respectively, $+69.3\%$ and $+42.5\%$ increase in Chamfer distance) and segmentation (respectively, $-15.8$ and $-17.4$ mIoU) results on average on all scenes of the \DATASETNAME. While these shapes are more \emph{compact} (fewer degrees of freedom), the associated reconstructions appear less legible and interpretable. They also fail to capture the diversity of real-world 3D data (houses, trees, windmills, boats, etc.).}
\section{Additional Qualitative Results}\label{sec:suppmat:quali}

\begin{figure*}[t]
    \centering
    \newcommand{\addarea}[4]{
    \rotatebox{90}{#1}
    & \includegraphics[trim={0 #4 0 #3},clip,width=.24\linewidth]{images/lidarhd/inputs/point_y/#2.jpg}
    & \includegraphics[trim={0 #4 0 #3},clip,width=.24\linewidth]{images/lidarhd/results/ours/point_y_pred/#2.jpg}
    & \includegraphics[trim={0 #4 0 #3},clip,width=.24\linewidth]{images/lidarhd/results/ours/rec_point_y/#2.jpg}
    & \raisebox{.5cm}{\includegraphics[width=.24\linewidth]{images/lidarhd/results/ours/allprotos/#2.jpg}}
    \\
}

\begin{tabular}{@{}l@{~}c@{~}c@{~}c@{~}c@{}}
    & Ground Truth & Semantic Segmentation & Reconstruction & Prototypes \\
    \addarea{~~\windturbine}{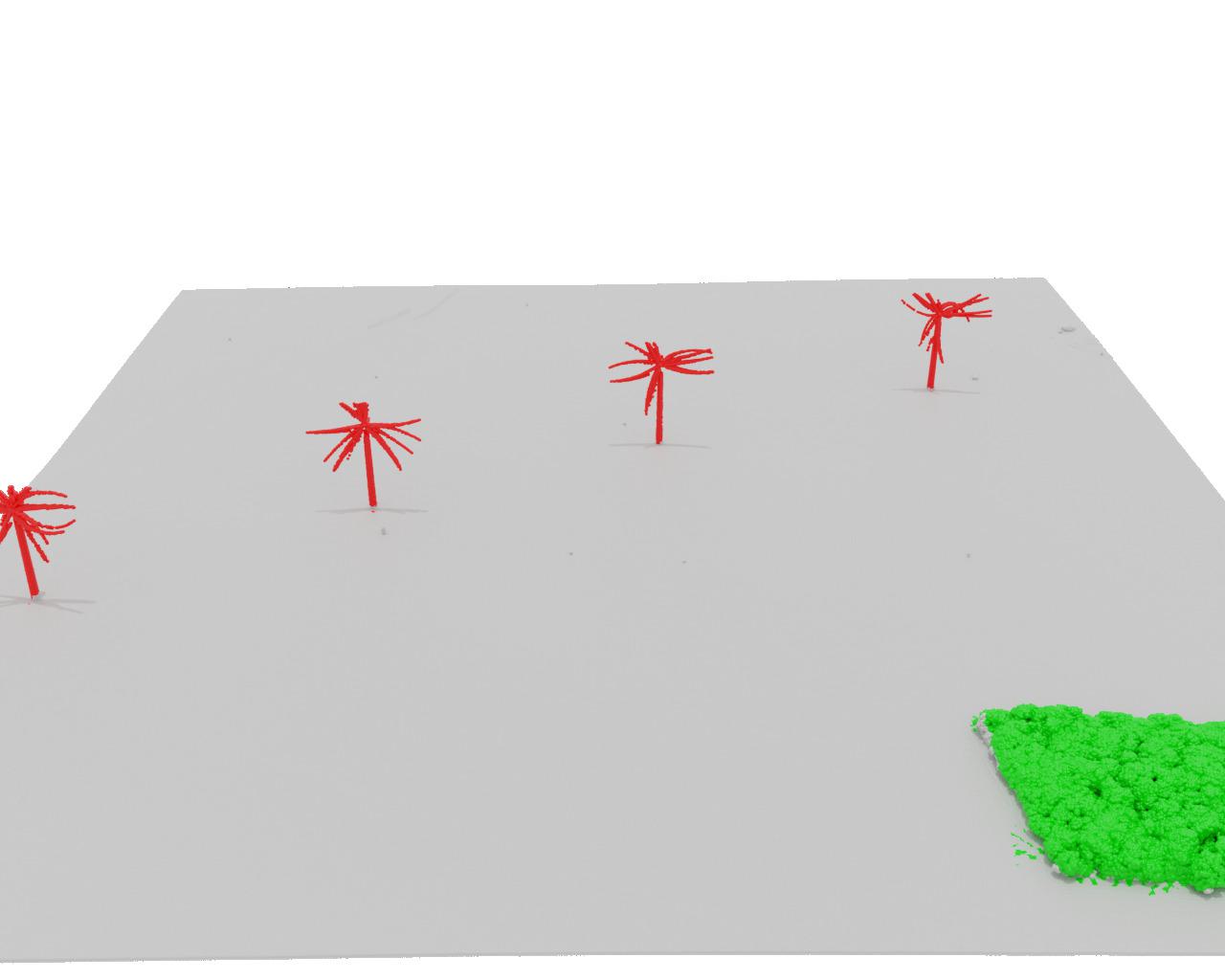}{200px}{0}
    \addarea{~~~~~~~~~\forest}{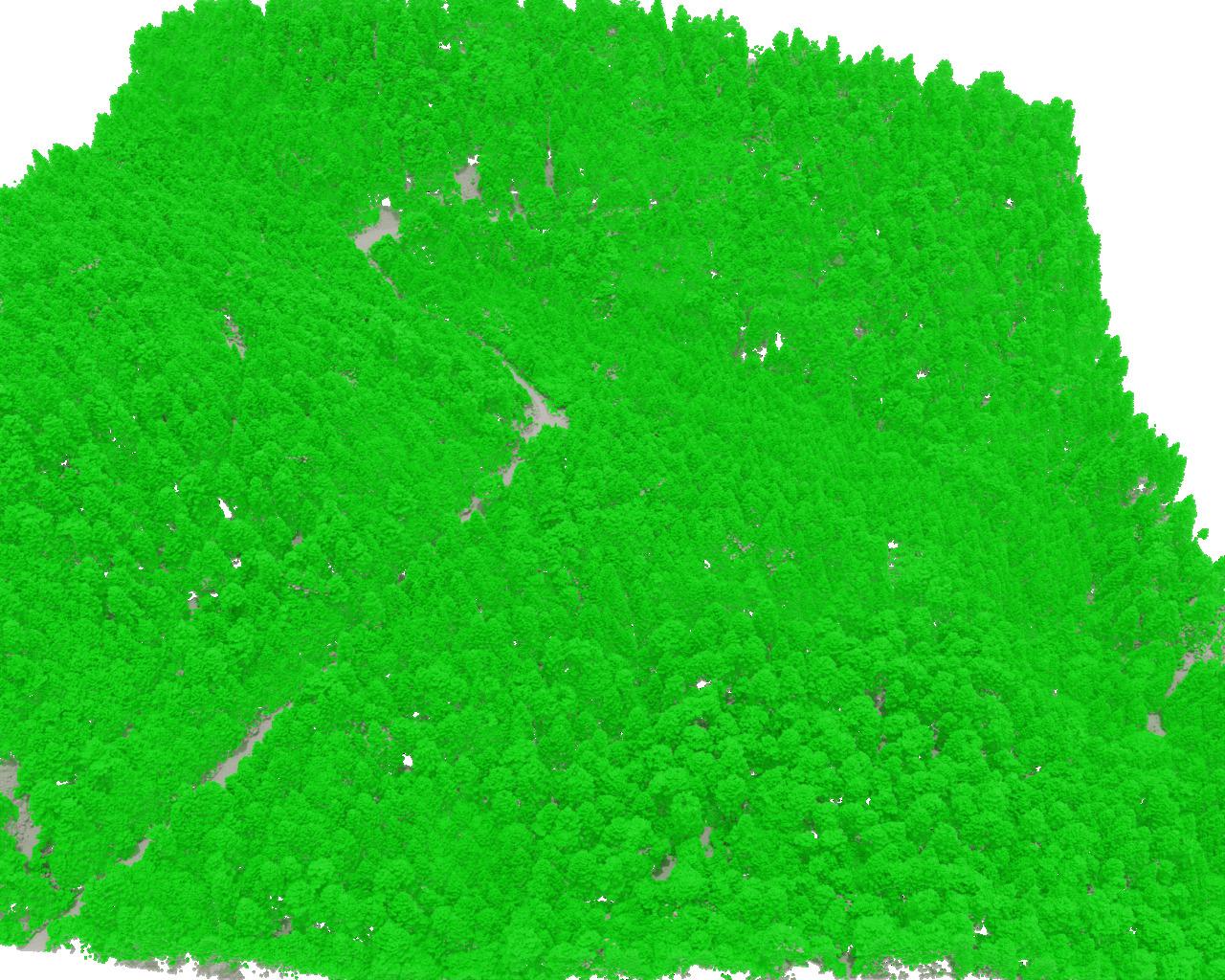}{0}{0}
    \addarea{~~~\cropfield}{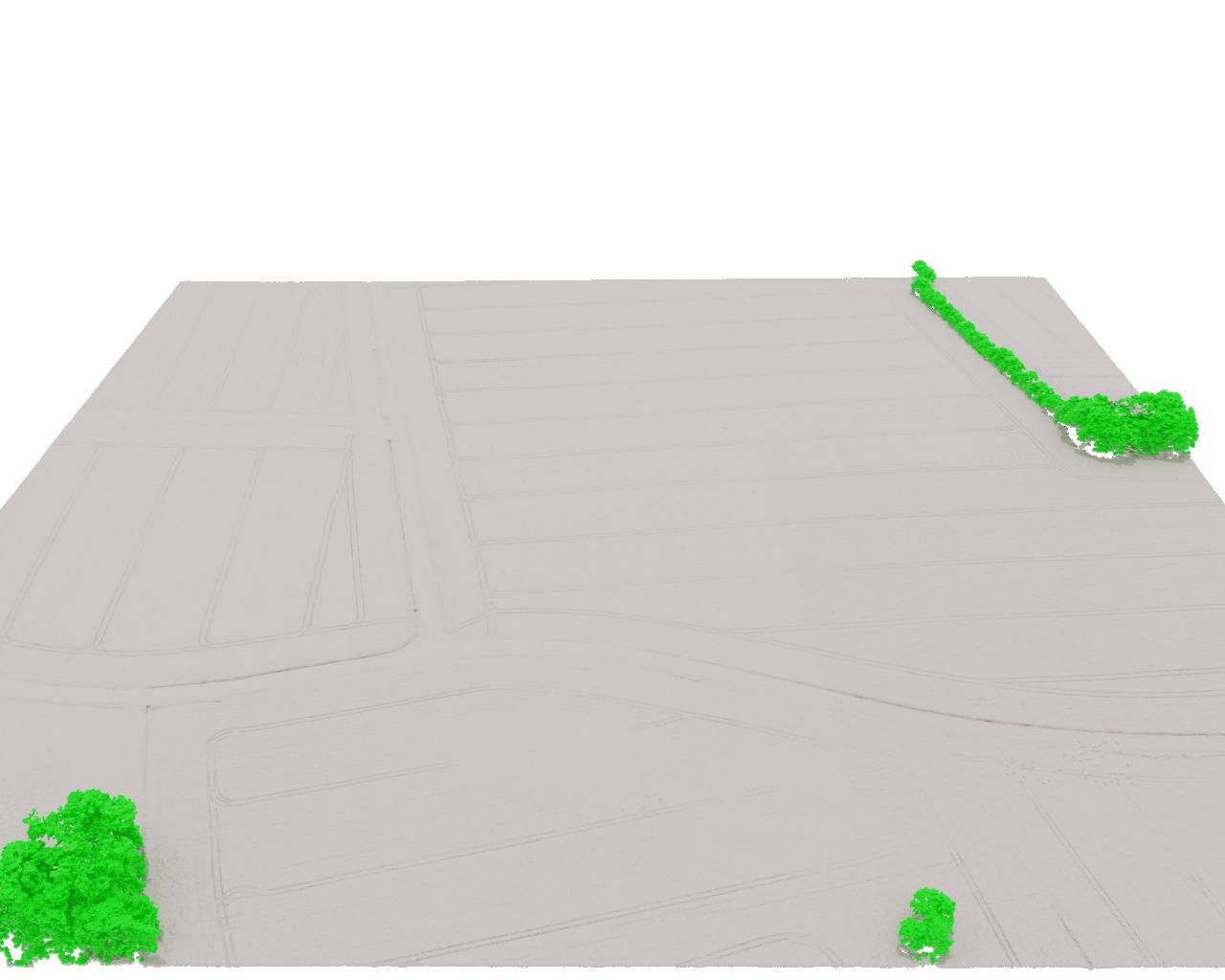}{200px}{0}
    \addarea{~~~~~~\nimes}{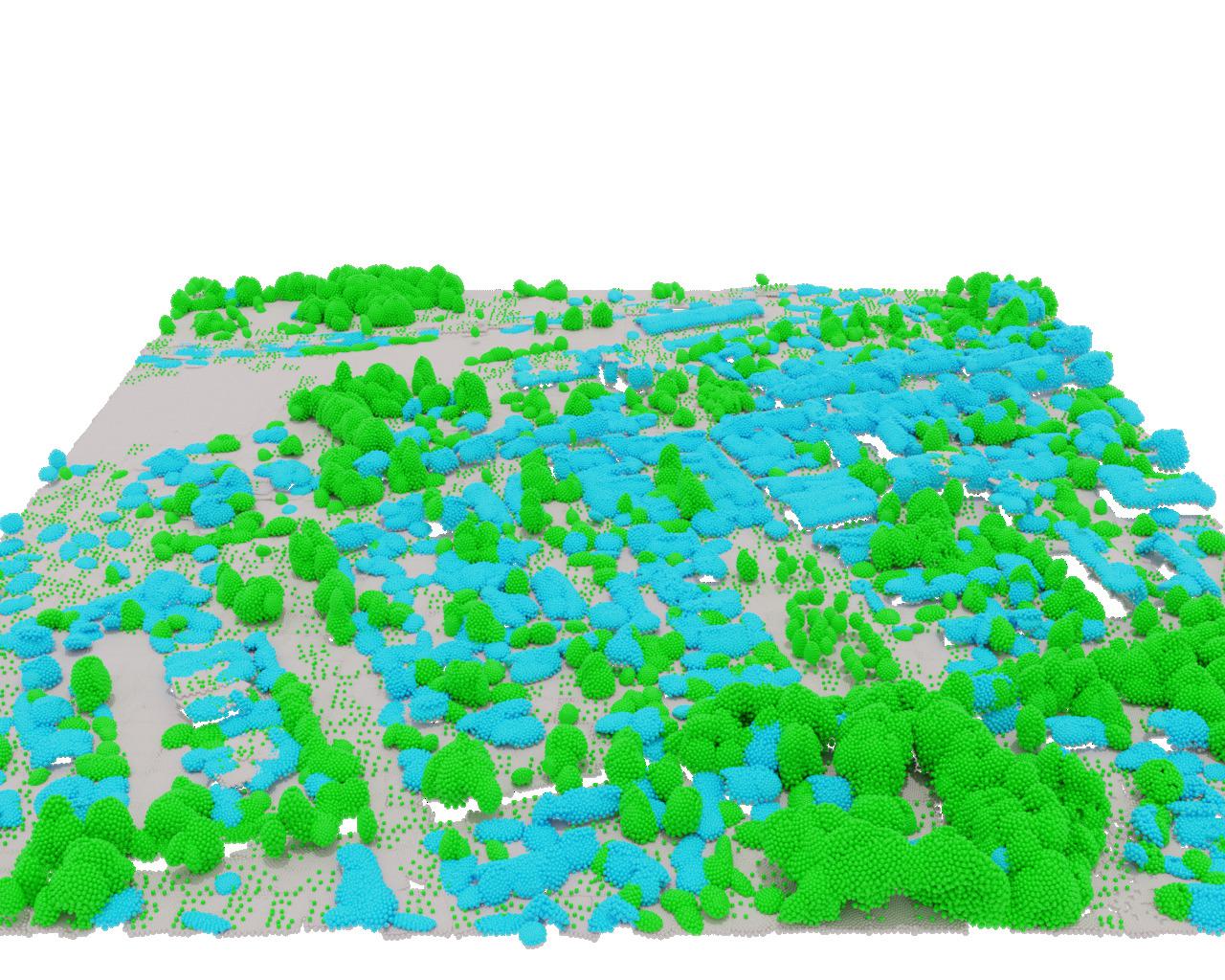}{200px}{0}
    \addarea{~~~\powerplant}{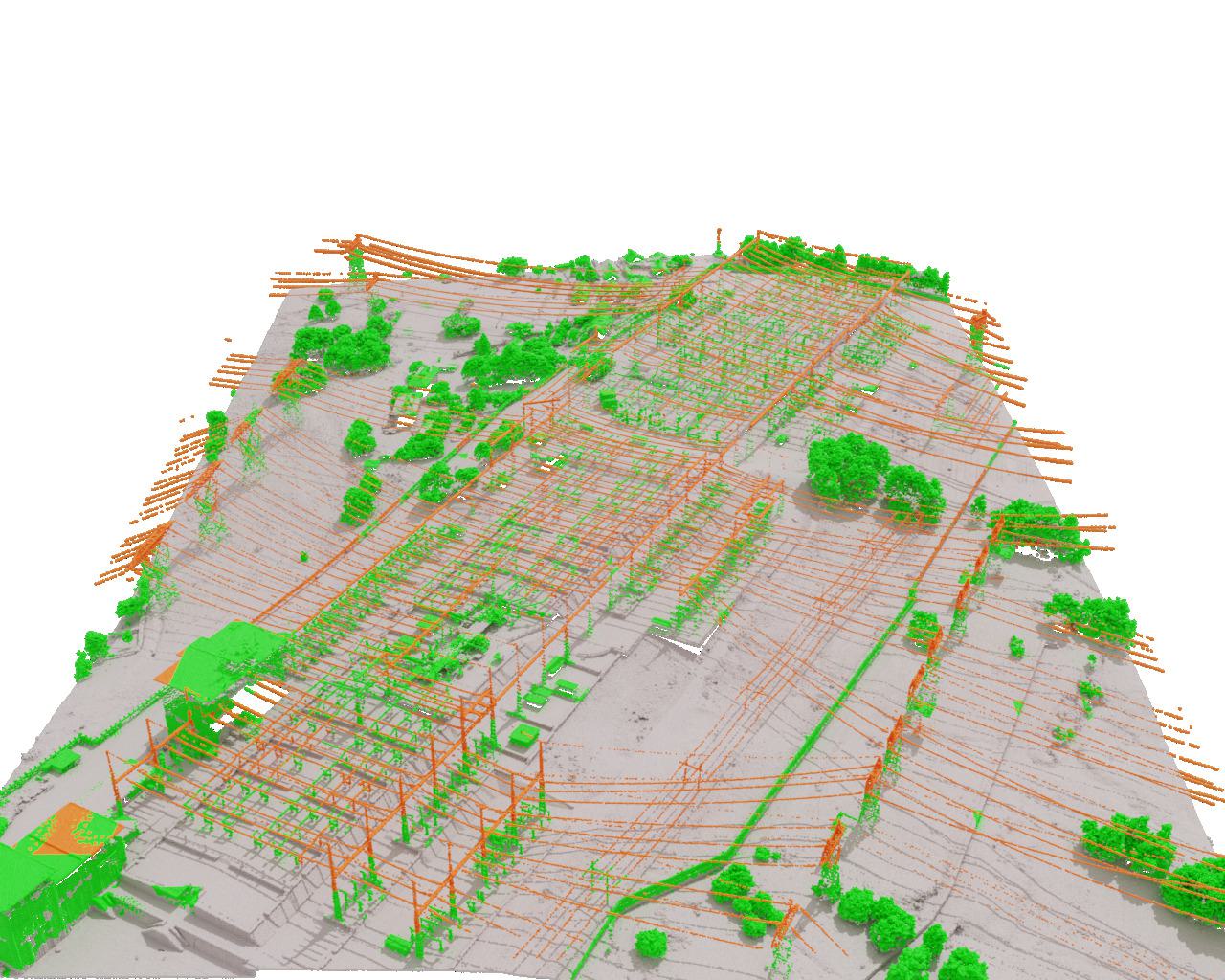}{160px}{0}
    \addarea{~~~\greenhouse}{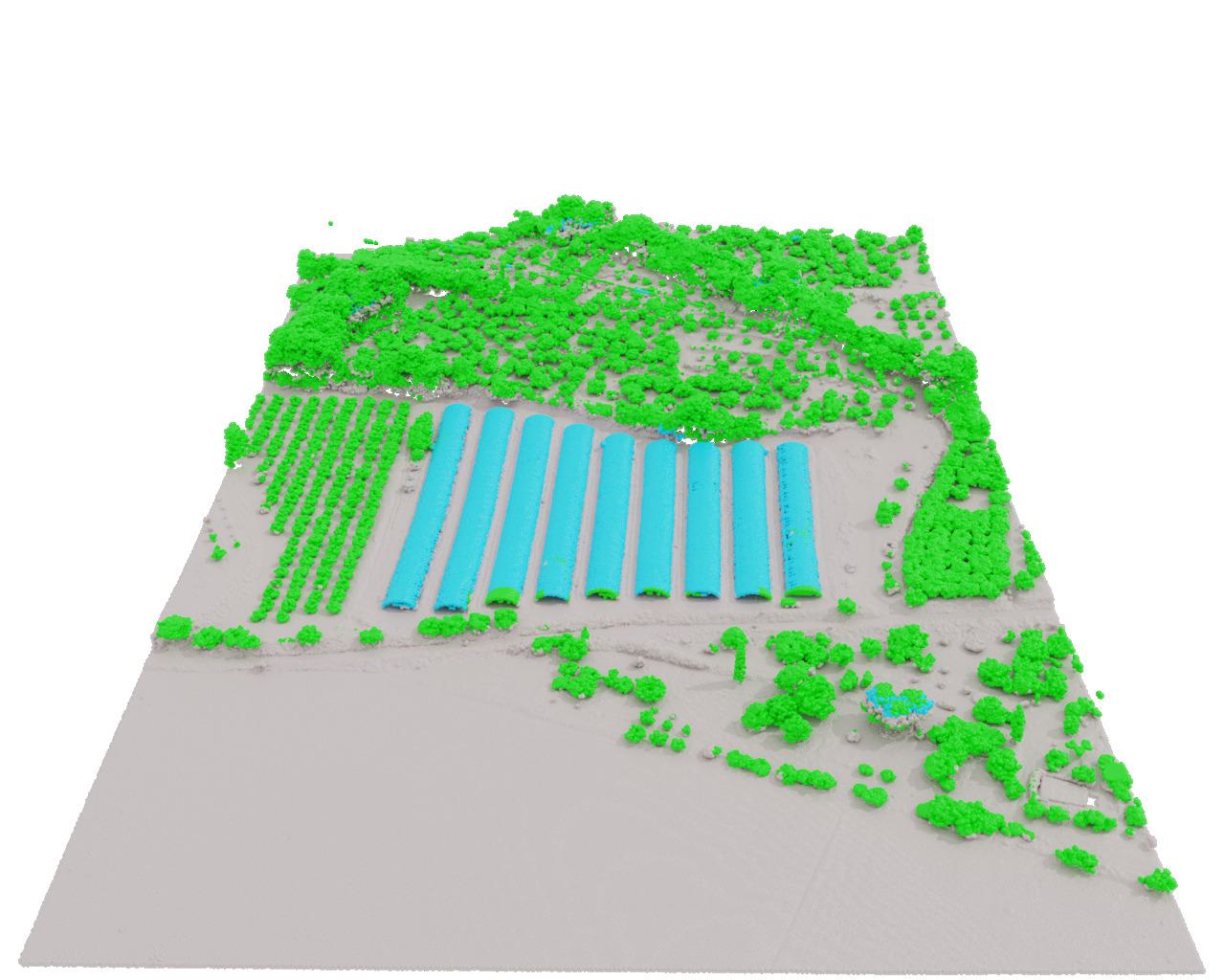}{160px}{0}
    \addarea{~~~~~~\boats}{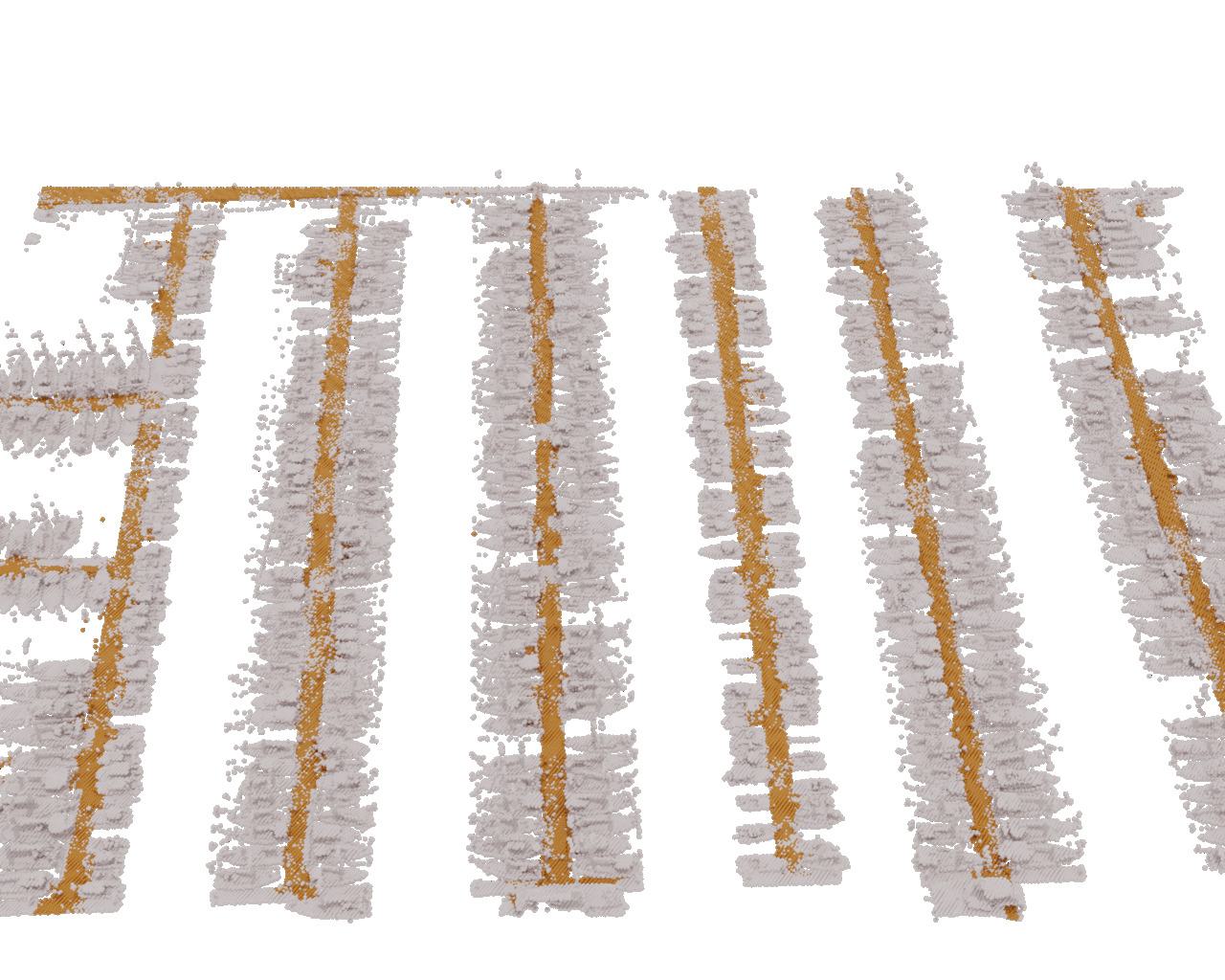}{150px}{50px}
\end{tabular}

    \caption{\textbf{Qualitative Results.}~For all scenes of the \DATASETNAME, we show the ground truth labels, the semantic segmentation, reconstruction, and prototypes learned by our \METHODNAME.}
    \label{fig:fullquali}
\end{figure*}

\begin{figure*}[t]
    \centering
    \setlength{\tabcolsep}{1pt}
\begin{tabular}{@{}ccc@{}}
    \includegraphics[width=.31\linewidth]{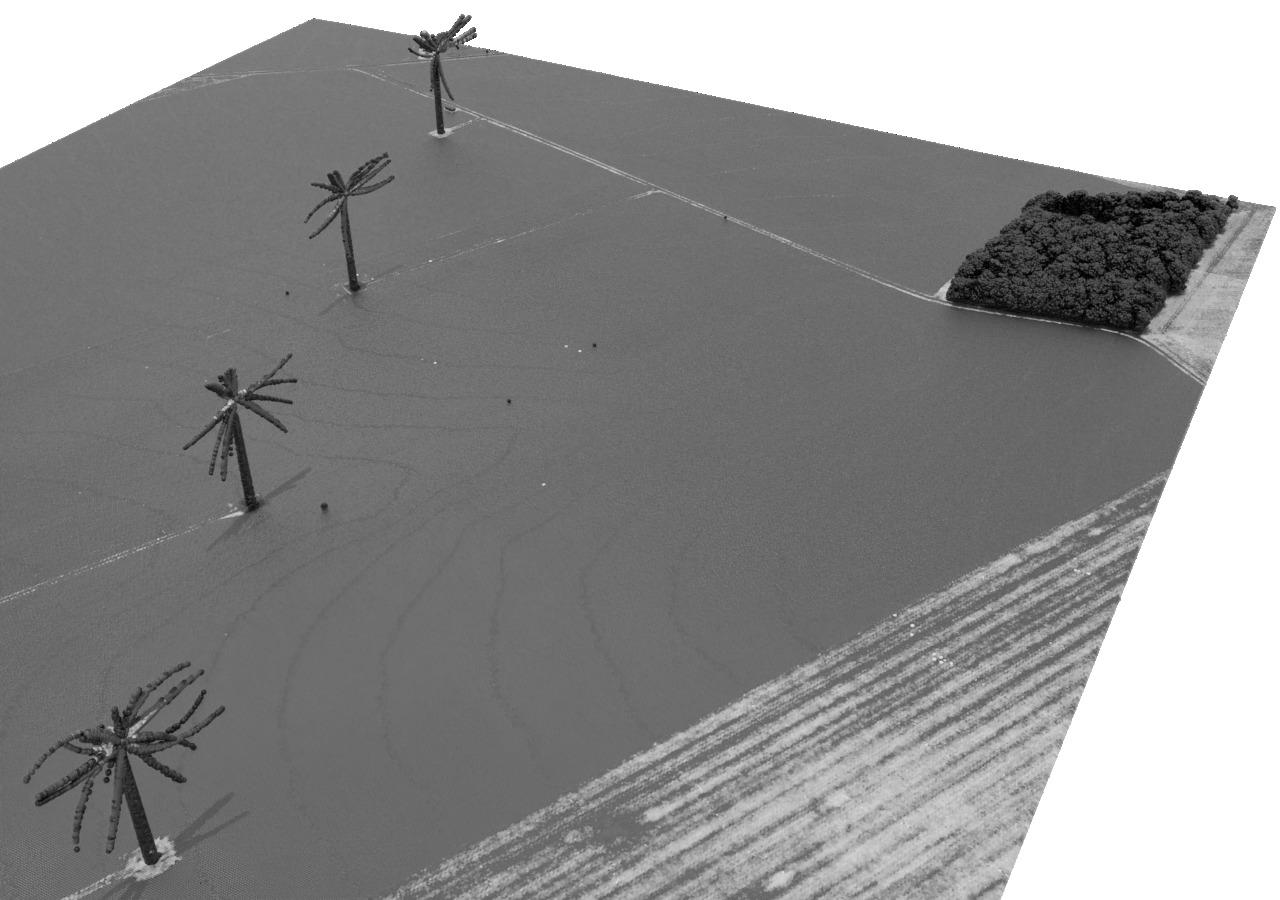}
    &
    \includegraphics[width=.31\linewidth]{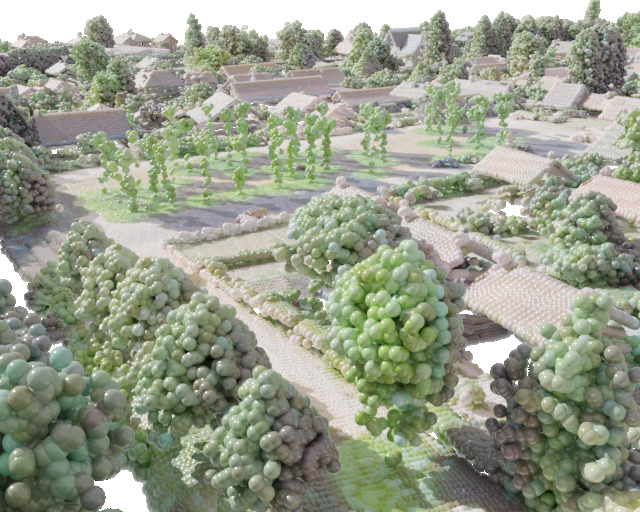}
    &
    \includegraphics[width=.31\linewidth]{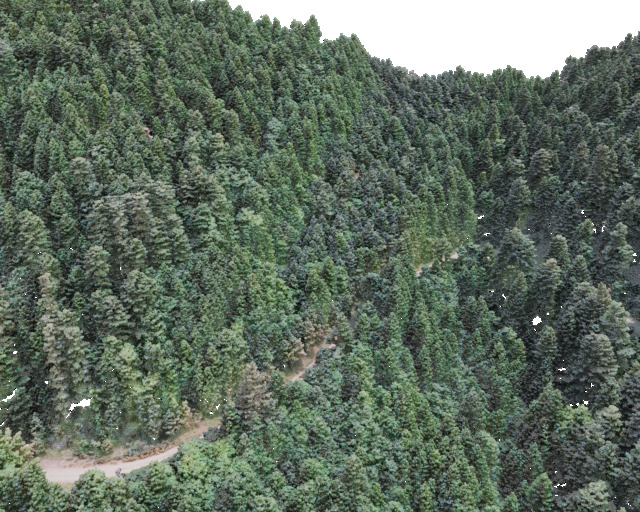}
    \\ 
    \includegraphics[width=.31\linewidth]{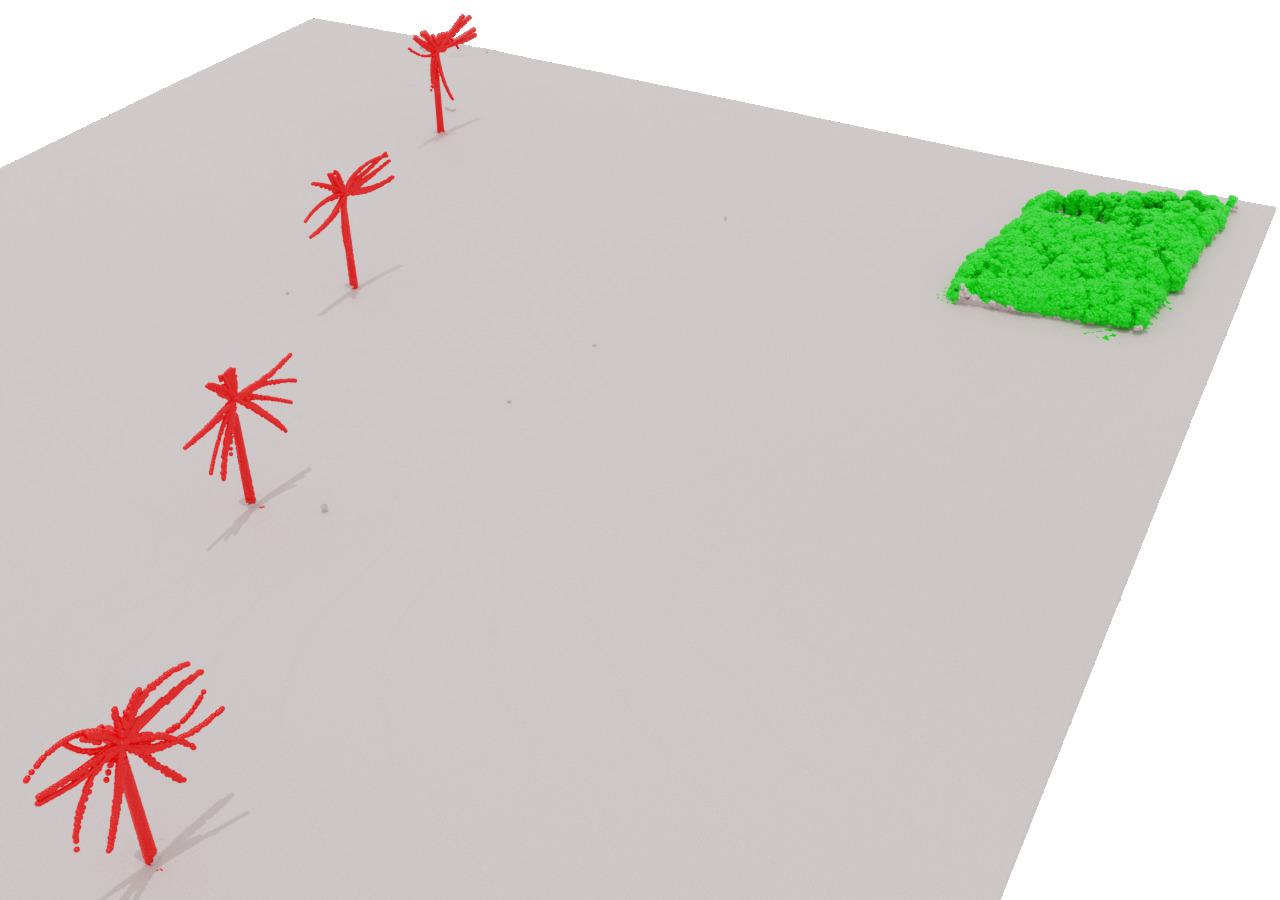}
    & 
    \includegraphics[width=.31\linewidth]{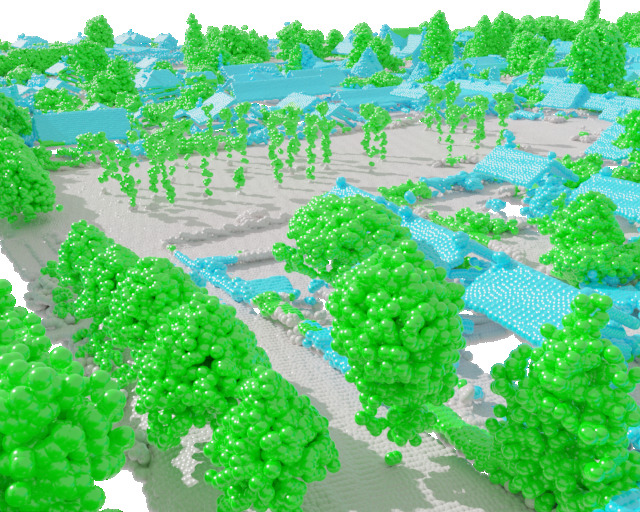}
    &
    \includegraphics[width=.31\linewidth]{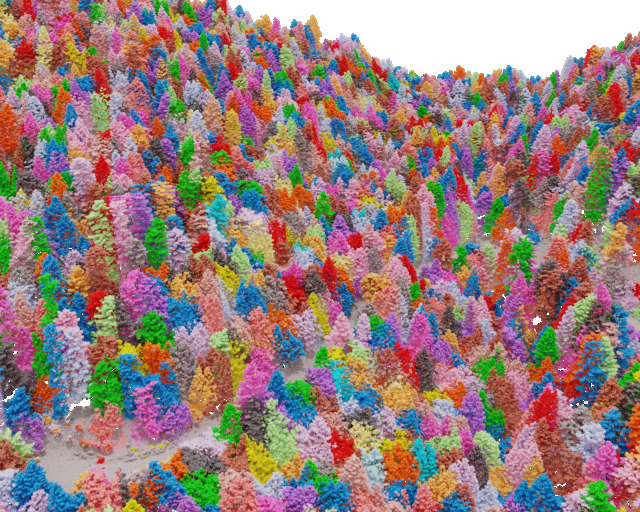}
\end{tabular}
    \caption{\textbf{Qualitative Semantic and Instance Segmentation.}~Our \METHODNAME~ can perform semantic and instance segmentation on various scenes with minor adaptations.}
    \label{fig:quali5}
\end{figure*}

\paragraph{\DATASETNAME~results.} We show in \figref{fig:fullquali} the ground truth semantic segmentation, our predicted semantic segmentation, our reconstruction and our learned prototypes. They showcase the quality, interpretability, and diversity of use cases of our model on this dataset of aerial LiDAR scans. We also show some semantic and instance segmentation closeups in \figref{fig:quali5}.

\begin{figure*}[t]
    \centering
    \centering
\setlength{\tabcolsep}{3pt}
\begin{tabular}{@{}c@{~}c@{~}c@{~}c@{~}c@{}}
    \rotatebox{90}{~~~~SuperQuadrics~\cite{paschalidou2019superquadrics}}
    &
    \includegraphics[width=.24\linewidth]{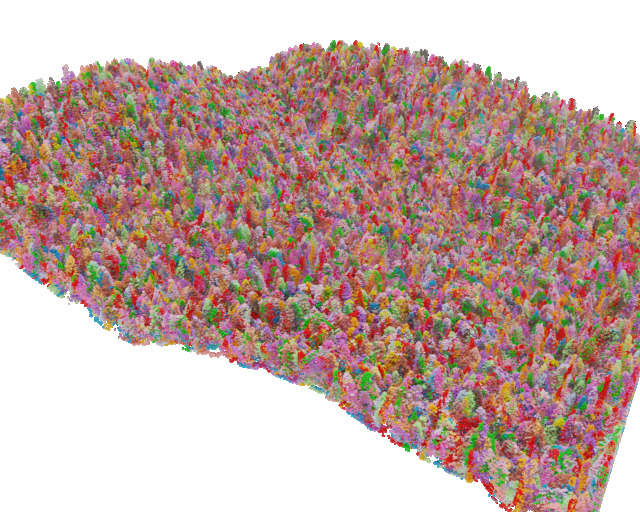}
    &
    \includegraphics[trim={0 0 5.5cm 4.5cm},clip,width=.24\linewidth]{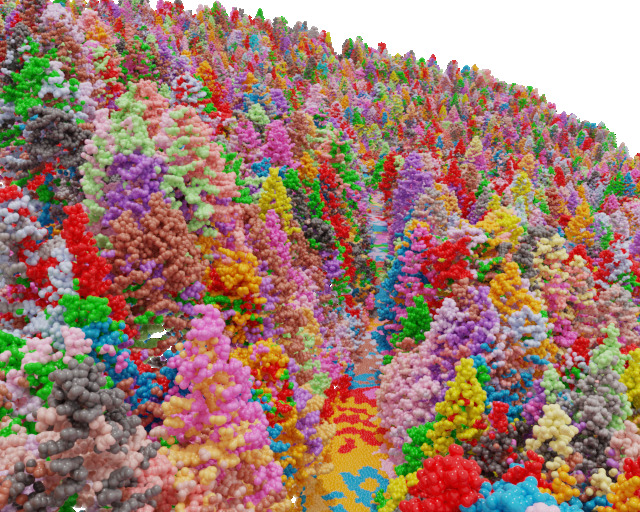}
    &
    \includegraphics[width=.24\linewidth]{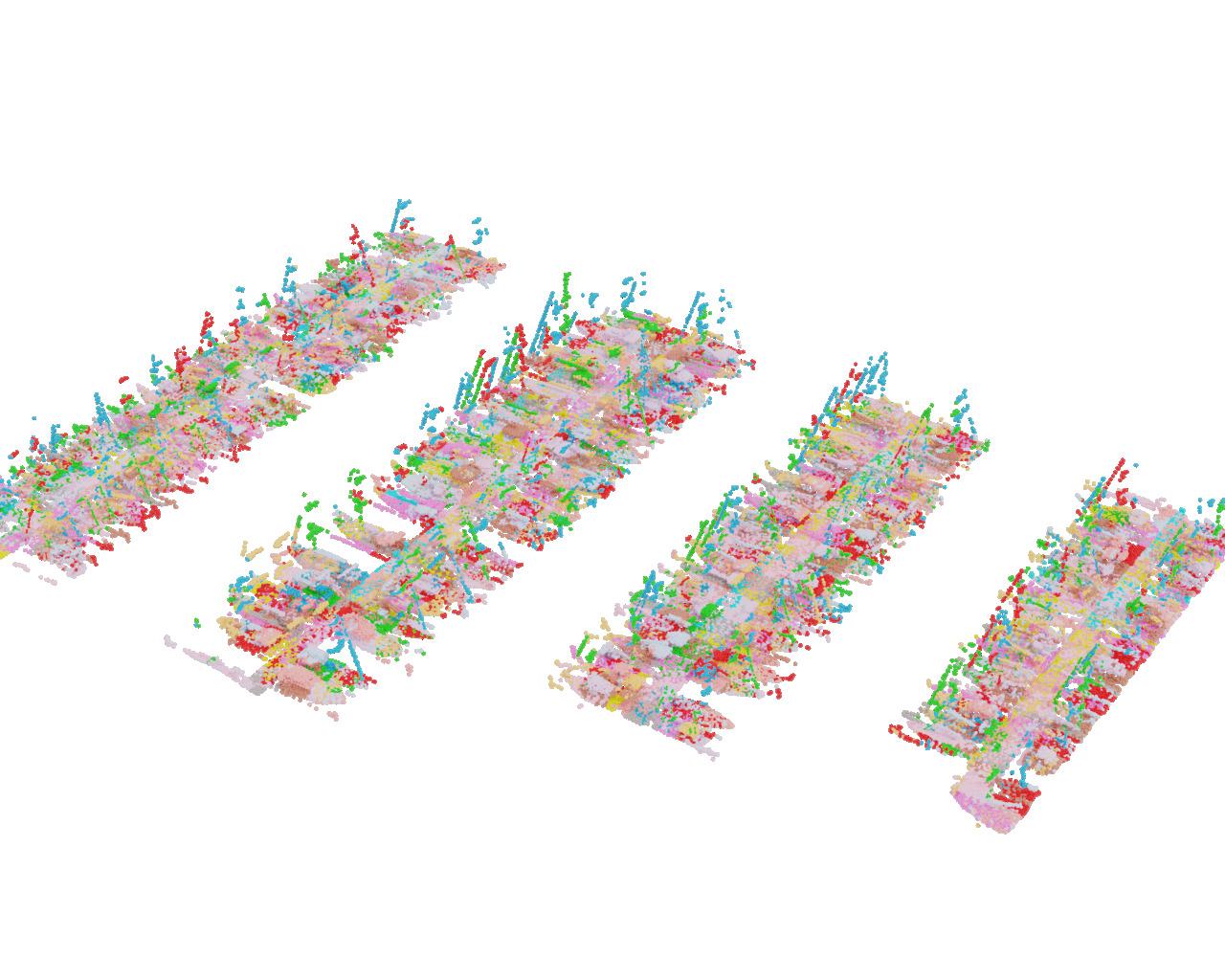}
    &
    \includegraphics[width=.24\linewidth]{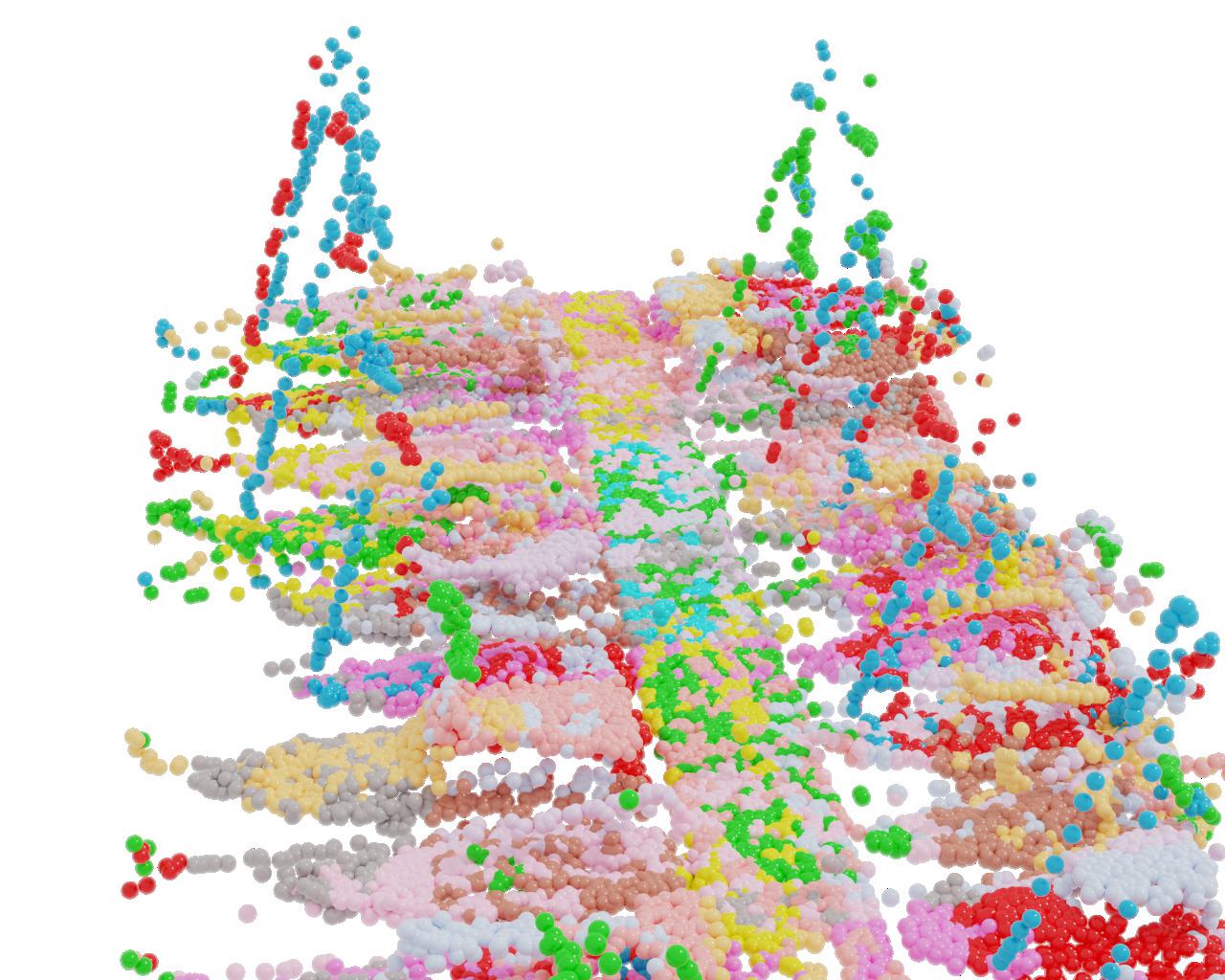}\\
    \rotatebox{90}{~~~~~~~~~~~~\textbf{Ours}}
    &
    \includegraphics[width=.24\linewidth]{images/zooms/forest_large_inst.jpg}
    &
    \includegraphics[trim={0 0 5.5cm 4.5cm},clip,width=.24\linewidth]{images/zooms/forest_zoom_inst.jpg}
    &
    \includegraphics[width=.24\linewidth]{images/zooms/large_boats_inst.jpg}
    &
    \includegraphics[width=.24\linewidth]{images/zooms/zoom_boats_inst.jpg}
\end{tabular}
    \caption{\textbf{Instance Segmentation.}~
    We represent the instances predicted with our algorithm and by SuperQuadric~\cite{paschalidou2019superquadrics}. We see that SuperQuadrics' reconstruction struggles modeling complex objects with only one instance. Moreover, our method make it easier to differentiate between different object types such as ``trees'' or ``boat hull'', while all superquadric are generated in the same way.
    }
    \label{fig:compinst}
\end{figure*}

\paragraph{Instance segmentation.}We show in \figref{fig:compinst} a comparison of the instance segmentation produced by SuperQuadrics~\cite{paschalidou2019superquadrics} and our \METHODNAME. Since SuperQuadrics~\cite{paschalidou2019superquadrics} uses a restricted family of 3D shapes to reconstruct an input scene, it has worst qualitative performances for instance segmentation when compared to our model, which learns scene-specific prototypes and can provide semantic information.

\begin{table*}[p]
    \centering
    \caption{\textbf{Results on the \DATASETNAME.}~We report the quality of the reconstruction (Cham.) and semantic segmentation (mIoU) for the models presented in the main paper and \textit{other variations}.
    \textbf{Bold} numbers indicate the best results of the models shown in the main submission, while \textbfg{green bold} numbers indicate the best results across all variations. \pdrop~indicates that the variation results in a significant drop in performance, while \pboost~indicates a performance boost. We also show the number of prototypes selected by our post-processing selection algorithm.}
    \resizebox{\textwidth}{!}{
\newcommand{\PreserveBackslash}[1]{\let\temp=\\#1\let\\=\temp}
\newcolumntype{C}[1]{>{\PreserveBackslash\centering}p{#1}}
\begin{tabular}{@{}lC{.9cm}C{.75cm}r@{~~}rr@{~~}rr@{~~}rr@{~~}rr@{~~}rr@{~~}rr@{~~}r@{}}\toprule
		       & \multirow{2}{*}{\rotatebox{45}{{Rec.}}} & \multirow{2}{*}{\rotatebox{45}{{Semantic}}} &\multicolumn{2}{c}{\cropfield} &\multicolumn{2}{c}{\forest}&\multicolumn{2}{c}{\greenhouse} 		&\multicolumn{2}{c}{\boats}		&\multicolumn{2}{c}{\powerplant}	&\multicolumn{2}{c}{\nimes}&\multicolumn{2}{c}{\windturbine}	\\\cmidrule(lr){4-5}\cmidrule(lr){6-7}\cmidrule(lr){8-9}\cmidrule(lr){10-11}\cmidrule(lr){12-13}\cmidrule(lr){14-15}\cmidrule(lr){16-17}
                           & & &Cham.	&mIoU	&Cham.	&mIoU	&Cham.	&mIoU	&Cham.	&mIoU	&Cham.	&mIoU	&Cham.	&mIoU   &Cham. & mIoU	\\\midrule
    k-means (i,z)~\cite{kmeans} & \xmark & \cmark	                                &---	&93.8	&---	&71.5	&---	&39.3	&---	&41.4	&---	&42.8	&---	&56.5	&--- & 87.6\\
\textit{k-means (i)}~\cite{kmeans}    & \xmark & \cmark                                 &---	&\pdrop74.5	&---	&\pdrop45.5	&---	&\pdrop36.3	&---	&\pdrop41.4	&---	&\pdrop28.8	&---	&\pdrop42.5   &--- &\pdrop64.1   \\
\textit{k-means (z)}~\cite{kmeans}& \xmark & \cmark                                     &---	&93.9	&---	&71.4	&---	&39.2	&---	&41.4	&---	&42.3	&---	&56.2   &--- &\pdrop77.5     \\\midrule
SuperQuadrics~\cite{paschalidou2019superquadrics} & 3D & \xmark & 0.86	&---	&1.04	&---	&0.60	&---	&0.93	&---	&0.58   &---	&0.40	&---	&13.5	&---\\\midrule
\DTI~\cite{monnier2021unsupervised} & 2.5D+i  & \cmark         &6.10	&83.2	&14.59	&40.2	&5.36	&42.0	&6.16	&41.4	&5.36	&29.0	&2.99	&47.3	&36.19	&25.9\\\midrule
AtlasNet~v2~\cite{deprelle2019learning}  & 3D+i & \cmark  &1.07	&43.1   &1.58	&71.4	&0.56	&49.1   &\bf0.73	&42.1	&0.45   &41.6	&0.63	&48.8   &9.47 & 48.1\\
\textit{AtlasNet~v2~\cite{deprelle2019learning} w/o intensity}  & 3D & \cmark   &1.08	&43.1   &\pdrop1.92	&\pboost74.4	&\pboost0.49	& \pdrop46.0   & \pdrop0.80	& \pdrop40.8	& 0.43   &\pdrop38.7	&\pdrop0.70	&\pdrop40.4   &\pboost7.56	& \pdrop25.9\\\midrule
\bf Ours
& 3D+i & \cmark     & \bf 0.72 & \bf 96.9 & \bf 0.88 & \bf83.7 & \bf0.40 & \bf 91.3 & 0.82 & \bf 78.7 & \bf0.44 & \bf52.2 & \bf0.29 & \bf83.2 & \bf6.65 & \bf93.4 \\
\bf \textit{Ours w/o post-processing}& 3D+i & \cmark                                        &\pdrop1.02	&96.5   &\pdrop0.97	&\pboost\textbfg{88.0}	&\textbfg{0.38}   &90.7   &\pdrop0.96	&78.3	&\textbfg{0.42}   &\textbfg{52.4}   &\textbfg{0.28}	&\textbfg{83.7}   &\textbfg{6.59} & \pboost\textbfg{96.4}	\\
\cdashline{1-17}
\textbf{Ours} \# of selected prototypes & --- & --- & \multicolumn{2}{c}{3} & \multicolumn{2}{c}{4} &\multicolumn{2}{c}{5} & \multicolumn{2}{c}{3} & \multicolumn{2}{c}{4} & \multicolumn{2}{c}{5} & \multicolumn{2}{c}{5}\\
\bottomrule\end{tabular}
}
    \label{tab:suppmat:quantidetails}
\end{table*}

\paragraph{Generalizability.} Our model trains individually per scene, taking around 12 hours each. We observed qualitatively that a model trained on one scene adapts well to other scenes of similar natures (e.g., different forests) but not otherwise. Training a universal model for diverse scenes is possible but would require significant memory due to the large number of prototypes needed.

\end{document}